\documentclass[review]{elsarticle}

\usepackage{hyperref}
\usepackage{todonotes}
\journal{Image and Vision Computing}

\begin{document}

\begin{frontmatter}

\title{Robustly Removing Deep Sea Lighting Effects\\ for Visual Mapping of Abyssal Plains}

\author[geomaraddress]{Kevin K\"oser\corref{mycorrespondingauthor}}
\cortext[mycorrespondingauthor]{Corresponding author}
\ead{kkoeser@geomar.de}

\author[geomaraddress]{Yifan Song}
\ead{ysong@geomar.de}

\author[geomaraddress,fhkieladdress]{Lasse Petersen}
\ead{lppetersen@geomar.de}

\author[geomaraddress]{Emanuel Wenzlaff}
\ead{ewenzlaff@geomar.de}

\author[fhkieladdress]{Felix Woelk}
\ead{felix.woelk@fh-kiel.de}

\address[geomaraddress]{GEOMAR Helmholtz Centre for Ocean Research Kiel, Wischhofstr. 1-3, 24148 Kiel}
\address[fhkieladdress]{Fachhochschule Kiel, Sokratesplatz 1, 24149 Kiel, Germany}

\begin{abstract}
The majority of Earth's surface lies deep in the oceans, where no surface light reaches. Robots diving down to great depths must bring light sources that create moving illumination patterns in the darkness, such that the same 3D point appears with different color in each image. On top, scattering and attenuation of light in the water makes images appear foggy and typically blueish, the degradation depending on each pixel's distance to its observed seafloor patch, on the local composition of the water and the relative poses and cones of the light sources. Consequently, visual mapping, including image matching and surface albedo estimation, severely suffers from the effects that co-moving light sources produce, and larger mosaic maps from photos are often dominated by lighting effects that obscure the actual seafloor structure.
In this contribution a practical approach to estimating and compensating these lighting effects on predominantly homogeneous, flat seafloor regions, as can be found in the Abyssal plains of our oceans, is presented. The method is essentially parameter-free and intended as a preprocessing step to facilitate visual mapping, but already produces convincing lighting artefact compensation up to a global white balance factor. It does not require to be trained beforehand on huge sets of annotated images, which are not available for the deep sea. Rather, we motivate our work by physical models of light propagation, perform robust statistics-based estimates of additive and multiplicative nuisances that avoid explicit parameters for light, camera, water or scene, discuss the breakdown point of the algorithms and show results on imagery captured by robots in several kilometer water depth.
\end{abstract}


\end{frontmatter}

%\linenumbers

% The first section title should be wrapped inside a \IEEEraisesectionheading as follows.
\section{Introduction}\label{sec:introduction}

\begin{figure}[t]
	\centering
	\includegraphics[width=0.5\columnwidth]{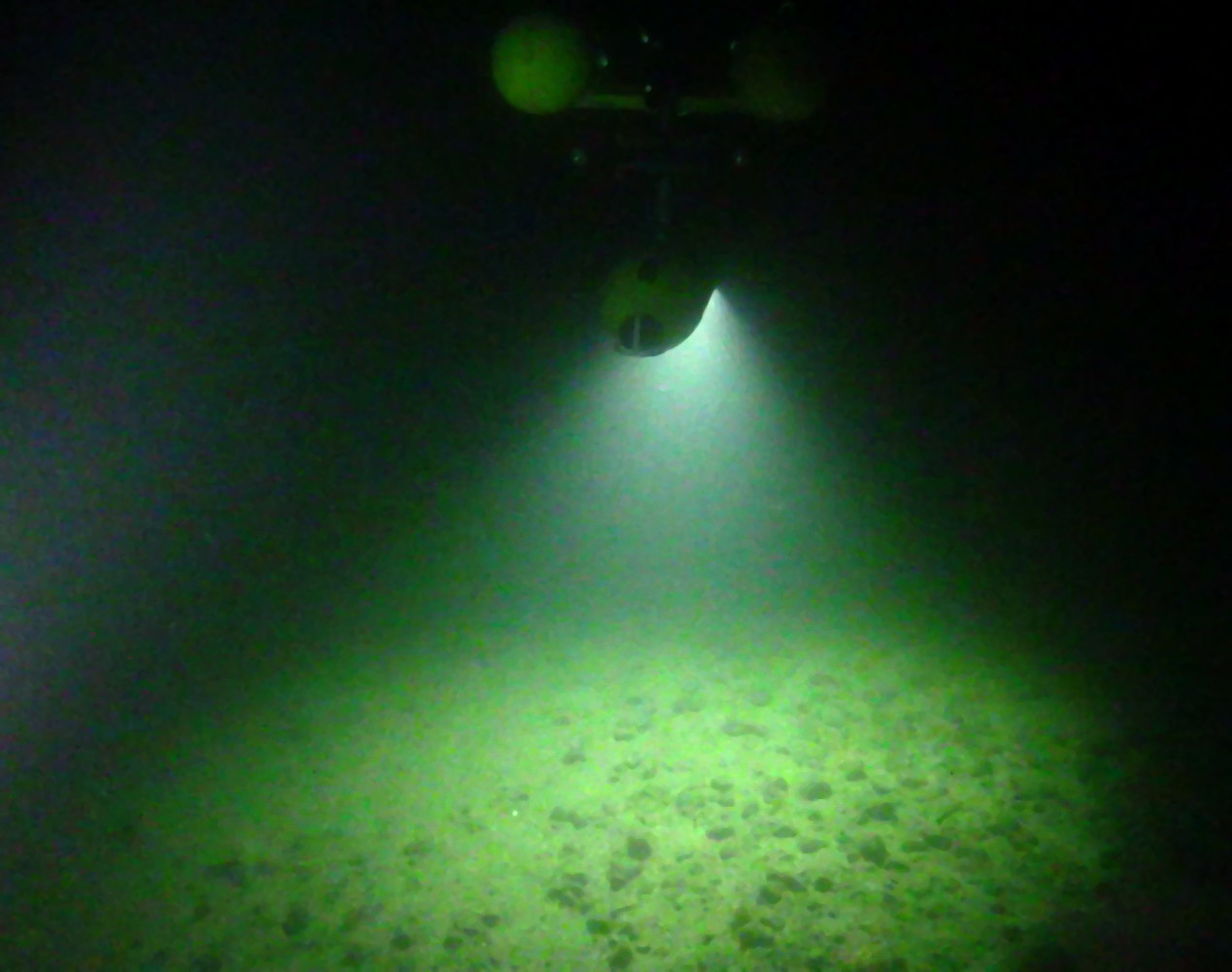} 
	\caption{An autonomous underwater vehicle (AUV) in about 100m water depth photographing the seafloor. Multiple LEDs with angular fall-off create illumination cones towards the seafloor. The footprint of the superimposed cones depends on the AUV's attitude and altitude, and the water body below the AUV scatters the light. Once the AUV moves on to the next photo, also the light will move on, making image matching and true surface color estimation very challenging.}
	\label{fig:teaseranton}
\end{figure}

% The very first letter of the paper is a 2 line initial drop letter
% followed by the rest of the first word in caps.
Although technology allows to map the surface of the Moon or even Mars, there are still large knowledge gaps for our own planet. More than half of Earth's surface is covered by at least one kilometer of sea water, and virtually all of this area has never been seen by any human and has not been visually mapped. The sunlight penetrates only a few hundred meters into the ocean and visibility of underwater cameras is limited to tens of meters under ideal (or laboratory) conditions and to typically less than 10m for deep diving robots in practice (even less in coastal waters, see fig. \ref{fig:teaseranton}). 

The lack of sunlight in the deep ocean requires robots to bring their own light sources, which creates two main problems: First, scattering of light can be viewed as a nuisance effect that makes images appear foggy, when light source and camera are relatively close to each other \cite{jaffe-90-optimalUnderwaterImagingSystems} (as the volumetric scattering function of different waters, e.g. measured in the seminal work of Petzold \cite{petzold1972volume}, has strong contributions into the "backwards" direction). Since a real deep ocean robot has to maneuver in a harsh environment and is deployed by a surface vessel, it has to obey to physical limits to size and camera-light layout. Robots that fly very close to the seafloor suffer less from scattering, but a robot at 5 meter altitude can cover way more area per hour as compared to a robot at one meter altitude (area footprint of an image grows quadratically with altitude, and cameras at higher altitude can move faster before motion blur becomes visible). Unless one is willing to use a team of robots with distributed light sources and cameras, image material taken by a single robot for large area mapping will be degraded by scattering. Second, when that robot is moving, also the light cones will move through the total darkness and create changing illumination at the seafloor (see e.g. \cite{song2021deepsea}). 
Additionally, distance-dependent absorption makes surface points appear darker or brighter (or change apparant color) when taking overlapping imagery of the seafloor. Even when traversing a flat seafloor at constant altitude, outer rays in a downward facing camera have traveled a longer way from the seafloor as compared to the central ray. The actual amount of scattering and attenuation depends on the (local) composition of the water and varies with wavelength and distance. Finally, rather than using an ideal point light source, a real robot uses real lamps, nowadays often multi-LED setups, that have an angular characteristic \cite{bryson16colorcorection,song2021deepsea}. Due to refraction effects, this characteristic can differ in air and in water. For energy budget reasons, and to achieve reasonably homogeneous illumination, often multiple light sources are mounted to the robot, and the pattern projected to the seafloor depends on altitude and the robot's 3D orientation. 

All these details make it really difficult to calibrate a lighting system and to determine all the physical parameters involved. However, the light and water effects significantly change the apparent color of a seafloor point when it is illuminated and observed multiple times from different viewpoints and distances, in particular for high-altitude mapping. These strong appearance changes impair image registration (both sparse and dense correspondence search) and light effects can dominate larger maps when not compensated. 

Classical underwater light propagation models \cite{McGlamery-1975-LightColModel,jaffe-90-optimalUnderwaterImagingSystems} have been used in the literature to undo some of these effects \cite{bryson16colorcorection}, but there are many parameters to estimate. On top, for scenarios, where we need to compensate strong illumination effects before we can estimate the motion, parametric physical models lead to a chicken-and-egg-problem, since the motion would be needed to compensate the light. When exploring never-before visited territories of our planet, methods that require huge amounts of training data are not appropriate either, and in general very little ground-truthed data exists for the deep sea, since it is practically impossible to see how the ocean floor would look without water.
%Recently, it has even been suggested that the classical (atmospheric) models using fixed attenuation and scattering coefficients for R,G and B might be insufficient to account for the underwater situation \cite{Akkaynak_2018_CVPR}, making the physical model even more complex.
 In this work, we therefore propose an empirical, parameter-free way of estimating illumination, attenuation and scattering, simply as multiplicative and additive terms that change the true surface color in the respective pixel. We show a robust but very simple way of estimating and compensating them, the overall method taking less than one second for a high resolution photo, allowing efficient processing of datasets of ten-thousands of high-resolution photos in a few hours and much quicker for lower (preview or live) resolution.

\section{Previous Work and Contribution}
Underwater imaging has a long history (see \cite{jaffe15underwaterimaging} for a recent overview). Tractable models for underwater lighting have been proposed by McGlamery\cite{McGlamery-1975-LightColModel} and Jaffe\cite{jaffe-90-optimalUnderwaterImagingSystems} and the low-level physics are discussed in detail by Preisendorfer and Mobley \cite{preisendorfer1964physical,mobley1994light}. Garcia et al. \cite{garcia02lighting} provide a general overview of lighting issues for robotics. In early work for post-processing after the dive, Pizarro and Singh\cite{pizarro03largeareamosaicing} divide each photo by the mean image of an entire mission, which imposes strong assumptions on altitude and attitude. As a parameter-free approach it is still used in ocean science practice\cite{morris_2014-autosub}, although it is not robust and does not account for scatter.

In general, two different types of approaches for tackling the underwater effects can be distinguished: Those that estimate the parameters of physical models and that undo the effects are called restoration techniques (e.g. \cite{schechner_05-underwaterPolarization,Trucco_2006-underwaterImgEnh,Sing_2007-towardsimaging,Treibitz_2009-PolarizationDescatter,Nicosevici_2009-efficientmosaicing,sedlazeck20093d,Williams_2012-benthicMonitoringAUV,Tsiotsios_2014_CVPR,bryson16colorcorection,Akkaynak19seethrough}). Though restoration methods have a solid physical interpretation, the models are often very complex and require perfect knowledge and calibration of many parameters such as position, orientation and angular characteristic of every light source, camera calibration, refractive interfaces of all lights and cameras, water absorption and scattering parameters, or complete distance information for each pixel in every image, which can be infeasible in practise.
Other, enhancement techniques, have been proposed that try to empirically improve image quality in fog, haze or underwater, e.g. by color histogram equalization, homomorphic filtering or using some assumptions about the scene (e.g. \cite{Bazeille_2006-imgPreProc,Iqbal_07-imgEnhancement,ancuti12enhancing,galdran18dehazingmulti-exposurefusion,santra2018transmittance,KIM2013410,zhao19multiscale}), see also \cite{Raimondo_2010-ColCorrectionStateOfArt,wang19reviewenhancement} for an overview. 
In contrast to restoration approaches, enhancement techniques usually do not require precalibrating all the parameters, but in particular single image enhancement methods are usually facing a strong ambiguity when trying to separate water effects, light cones and surface texture. Predicting plausible heuristics for previously unvisited deep sea territories is challenging. Also mixtures of pure empirical and strict physical models have been developed. 
These typically estimate a depth map from a single image (e.g. like using the dark channel prior in air \cite{he11darkchannel}) and then use this approximate geometric layout to invert a parametric underwater imaging model \cite{ancuti16descattering,peng15blurriness,UnderwaterHazeLines}. Learning water and illumination effects as for shallow water\cite{Li_2017} is difficult, because very little training data exists for how the ocean floor would look without water and manually correcting images is infeasible for human annotators. 

Despite the huge number of listed approaches above, almost all of them are designed for shallow water with sunlight and most deal with variants of the fog model\cite{cozman1997depth}.
This an important setting in coastal areas and for diver scenarios in the top few meter of the ocean, but the lighting regime is entirely different from the deep sea\cite{song2021deepsea}. 
Only \cite{Sing_2007-towardsimaging} and \cite{bryson16colorcorection} consider light cones and the deep sea scenario, although none of the two compensates additive backscatter. In an approach inspired by homomorphic filtering, Singh et al.\cite{Sing_2007-towardsimaging}, fit a 4th order polynomial to an image in log space in order to represent the multiplicative illumination. This can capture the effects of a single light cone, but will also vary depending on the seafloor structures, i.e. is image dependent. On top, for multi-LED setups the degree of the polynomial fitted has to be adapted to the complexity of the illumination pattern. The work of Bryson et al.\cite{bryson16colorcorection} on the other hand requires to model and calibrate each of the light sources jointly with the camera in order to undo the lighting effects. While this is a desirable solution in theory, obtaining all light source, camera and water parameters for a heavy deep sea robot with 24 LEDs can be challenging. On top, as we argue in this paper, robot localization and mapping can benefit from previous image enhancement, which is however not possible in case the enhancement itself already requires the results of the robot localization and mapping (chicken-and-egg problem).

Consequently, in this paper we propose a new calibration-free method that robustifies the mean image idea of \cite{pizarro03largeareamosaicing}, extend it to scenarios with significant backscatter and generalize to missions with varying altitude/attitude. At the same time we analyse assumptions, applicability and breakdown point in detail. The novel contributions and desirable properties are as follows:
 
 % follow this method and use a narrow field of view camera (27$^\circ$) at relatively low altitudes to photograph only the central, homogeneously illuminated region of the seafloor (2.4m$^2$ per picture, whereas we aim at 100m$^2$-200m$^2$ per picture).

% but the majority of the underwater vision mapping literature focuses on shallow water in sunlight (e.g. \cite{ }), which 
 
% Despite varying altitudes they compute one global illumination mask by taking the mean of 50 (presumably hand-picked) seafloor images, and then tackle the water effects using a physical model and ocean parameters from the literature.
 
% Essentially, 
 %  We build on top of the first part of this approach with the following novel contributions and desirable properties:
\begin{enumerate}
	\item We show that effects can be categorized into additive or multiplicative nature. We then perform automatic, {\em robust} estimates of the sum of the additive and the product of the multiplicative components. These estimates do not suffer from floating particles or occasional bright or dark seafloor patches and no user interaction is needed.
	\item We give clear preconditions in what scenarios the algorithms will work (breakdown point of robust estimator to observe the dominant seafloor). The only steering parameter (filter size) is rigorously derived from seafloor properties (percentage of uniform seafloor).
	\item Rather than assuming a fixed light pattern at the seafloor, we only assume the additive component (scatter) to be static during a mission and dynamically re-compute the multiplicative estimate per image, allowing us to cope with varying altitudes and changes in vehicle orientation.
	\item The approach does not require calibration of physical parameters, and we do not require any knowledge about light orientation and distance, nor lens calibration, nor water properties, nor a 3D model of the scene, making the approach attractive even for old videos with unknown parameters.
	\item The sliding-window techniques are compatible with a streaming architecture allowing for a (near) real-time implementation for estimation and compensation.
	\item Since obtaining ground truth for deep waters is close to impossible, we propose a new objective metric for computing the restoration quality on real data without ground truth: Overlapping imagery should restore the same color for corresponding pixels. The proposed metric does not require the true color and avoids a bias towards dark or low contrast restorations.	
\end{enumerate}

We also explicitely re-sketch the artificial light and water effects for the deep sea scenario with co-moving light sources (see also \cite{jaffe-90-optimalUnderwaterImagingSystems,song2021deepsea}). This is not a new derivation, but we believe that it is important for readers to distinguish this scenario from shallow water settings with sunlight often approximated by the ``fog model``\cite{cozman1997depth}.
%Though this is not a novel contribution, we believe that this is important for readers to distinguish the scenario from the shallow water setting with sunlight and needs an entirely different lighting model. For instance, as exemplified for Jerlov water type II, most of the scattering visible originates from close to the camera.

Once enhanced or restored versions of the original images have been created, usually some water or lighting effects remain. In order to create large maps, different mosaicing and blending strategies can be applied (see \cite{prados11blending} for a discussion). Note however that the main goal of blending strategies is to make artefacts less prominent (e.g. by distributing intensity discrepancies over a larger area rather than creating a hard edge), or to create visually pleasant maps. Our goal is to improve the consistency of the input images, such that they can be used for visual mapping purposes (feature correspondences, dense reconstruction, loop detection and texturing).

\section{Parameter-free Light Compensation}
\begin{figure}[t]
	\centering
	\includegraphics[width=0.48\columnwidth]{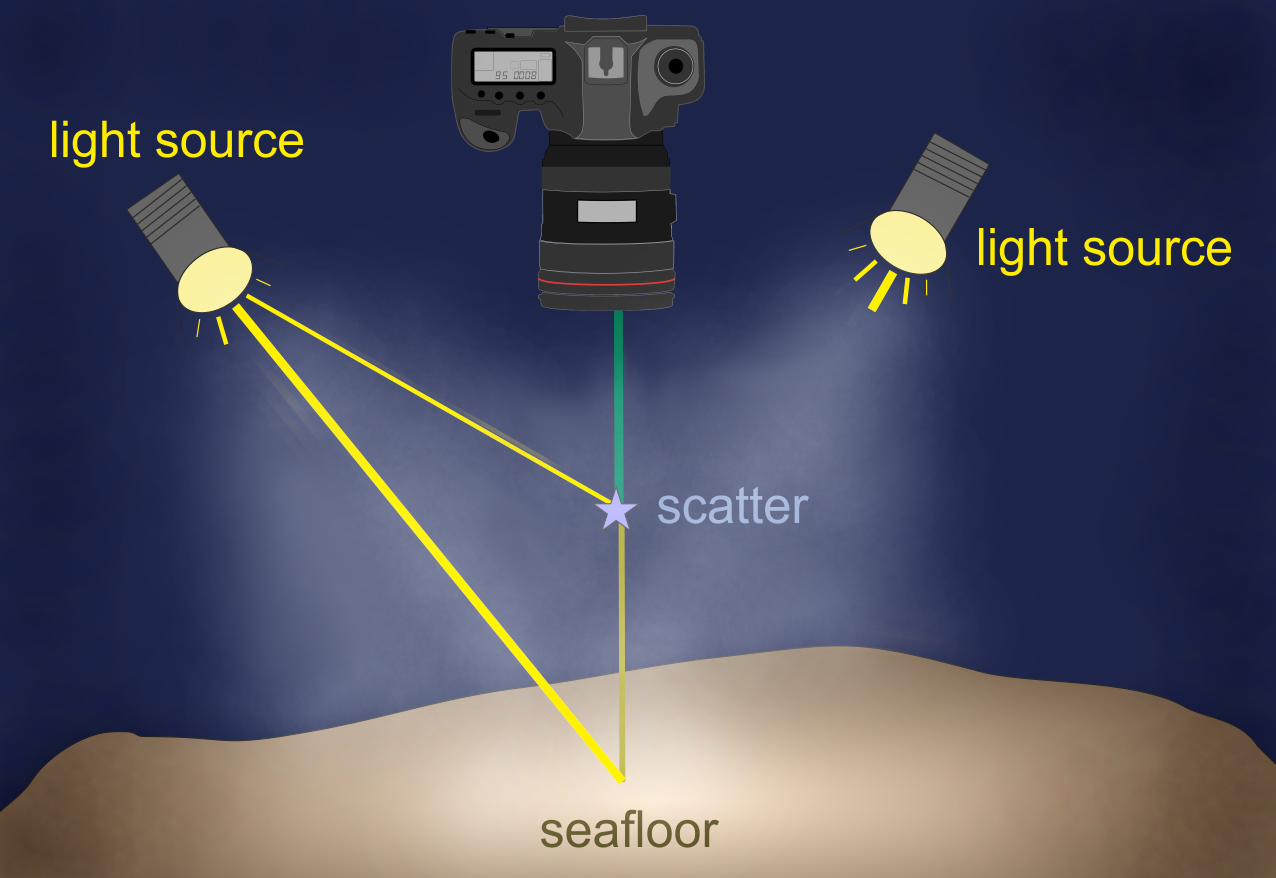} 
	\caption{In the deep sea scenario all the light originates from the light sources that rigidly move with the camera. Depending on the vehicle design, multiple light sources with individual positions and orientations are used. Each light source could have its own characteristic angular emission pattern and different regions of the ground receive different irradiances as the superposition of water-attenuated light from all light sources. This incoming light (spectrum) is then basically multiplied by the albedo (spectrum) of the ground, assuming a Lambertian surface. On its way to a pixel of the camera chip, the signal from the ground is superimposed (additive) by other photons that are scattered into the optical corridor of the same pixel. Compare also \cite{McGlamery-1975-LightColModel,jaffe-90-optimalUnderwaterImagingSystems}}
	\label{fig:lightsketch}
\end{figure}
\begin{figure}[t]
	\centering
	\includegraphics[width=0.48\columnwidth]{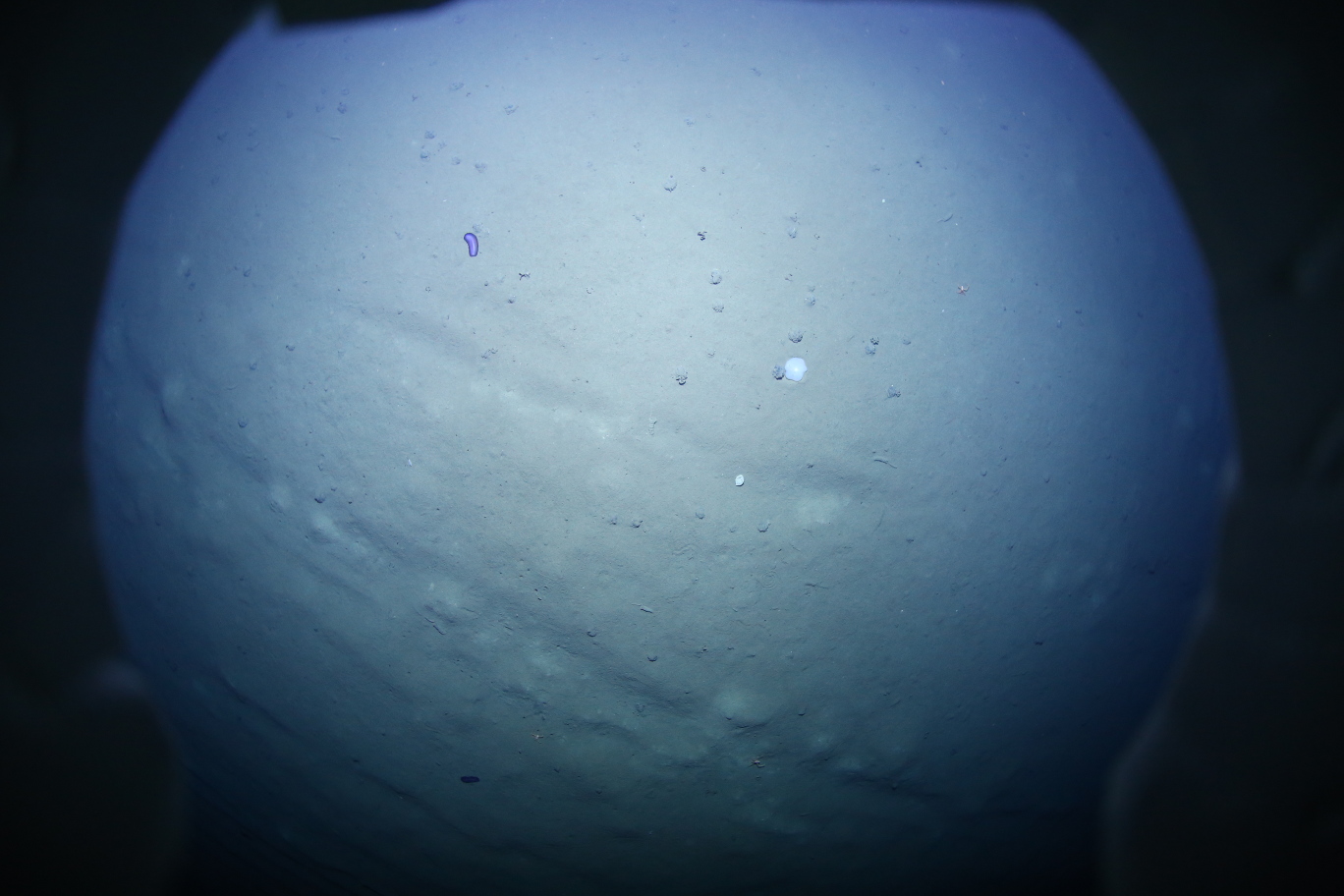}
	\caption{A seafloor image taken by a deep sea AUV in about 4km depth, using a full format DSLR with a 15mm fisheye lens.}
	\label{fig:rawimage}
\end{figure}

To illustrate the assumptions of the model used, we will look into figure \ref{fig:lightsketch} and inspect a particular ray that reaches a sensor pixel (and the sensor pixel will integrate over a range of wavelength and a range of spatial rays). Basically, the incoming light along a ray originates from a non-uniform point light source with angular characteristics $I_l(\theta,\phi)$. Commonly, several or many light sources are used, and the light from each of them has to be considered, but for clarity, we will just mention one light source here (since there is no interaction of light sources, the light received from a multi-light system is just the sum of all individual lights). The directional pattern of a light source might be quite complex and so in this contribution we do not attempt to estimate it, but use the term $I_l(\theta,\phi)$ just for illustration purposes. Following one of the directions from the source, a fraction of the light will hit the seafloor after having traveled the lightsource-seafloor distance $d_l$, and the light is attenuated along this distance. The seafloor reflects some amount $A_s$ of the incoming light according to its bidirectional reflectance distribution function towards the direction of the camera (for Lambertian surfaces, $A_s$ can be considered the color, or the albedo, of the surface, weighted by the cosine of the incident illumination). The reflected light then travels a distance of $d_c$ from the seafloor to the camera and is attenuated along the way, before some of this light reaches the image sensor. On top, also non-desired light is back-scattered into the optical path, and the sum of all scattering $I_s$ along the ray adds to the previously explained light component. So the overall intensity $I_c$ received at a camera pixel can be modeled as
\begin{equation}
\underbrace{I_c}_{\mathrm{observation} }  = \underbrace{I_l(\theta,\phi)}_{\mathrm{light \; source}} \cdot \underbrace{e_{\;}^{-\eta d_l}}_{\mathrm{atten.}} \cdot \underbrace{A_s}_{\mathrm{seafloor}} \cdot \underbrace{e^{-\eta d_c}_{\;}}_{\mathrm{atten.}} + \underbrace{I_s}_{\mathrm{backscatter}} ,
\label{eq:lighting}
\end{equation}
where $\eta$ is a water-specific attenuation coefficient.
Note that $I_l$, $\eta$,$A_s$, $I_l$ are all wavelength dependent. Since each pixel sees a different seafloor point, $\theta$ and $\phi$ as well as the distances $d_l$ and $d_c$ will vary with pixel position $x$, and also the backscatter  $I_s$ "collected" along the respective line of sight depends on the pixel position. See fig. \ref{fig:rawimage} for an example image.

If all parameters, including the camera/light pose dependent distances and angles, were perfectly known, one could try to use image restoration techniques to solve for the seafloor albedo $A_s$, although this is a challenging problem already in shallow water \cite{Akkaynak19seethrough}.
However, prior to optical localization and mapping, distances and orientations of camera and lights to each seafloor point are not known exactly, or for old videos important parameters might be missing or detailed localization and mapping be infeasible. Even when planning a mission, geometric and radiometric calibration of a multi-light and camera system is a challenging task. Consequently, in the following we describe how the numerous individual parameters can be grouped into larger ``combined effects`` which can be obtained from the statistics of the acquired images, if there is a predominant seafloor color (sediment, sand, etc.). Essentially eq. \ref{eq:lighting} can be rearranged into multiplicative and additive effects, and we will write the product of all factors as a function $F$:
\begin{equation}
I_c = A_s \cdot F(\phi, \theta, d_l, d_c, \eta) + I_s,
\label{eq:simplifieda}
\end{equation}
Actually, $\phi, \theta, d_l, d_c$ all depend on the pixel position $x$ in the image and on the relative pose $P$ between the vehicle and the ground ($\eta$ only depends on the wavelength we are considering). To make this dependence more clear, we explicitly write
\begin{equation}
I_c(x,P) = A_s(x,P) \cdot F(x,P) + I_s(x,P),
\label{eq:simplified}
\end{equation}
\begin{figure}[ht]
	\centering
	\includegraphics[width=0.48\columnwidth]{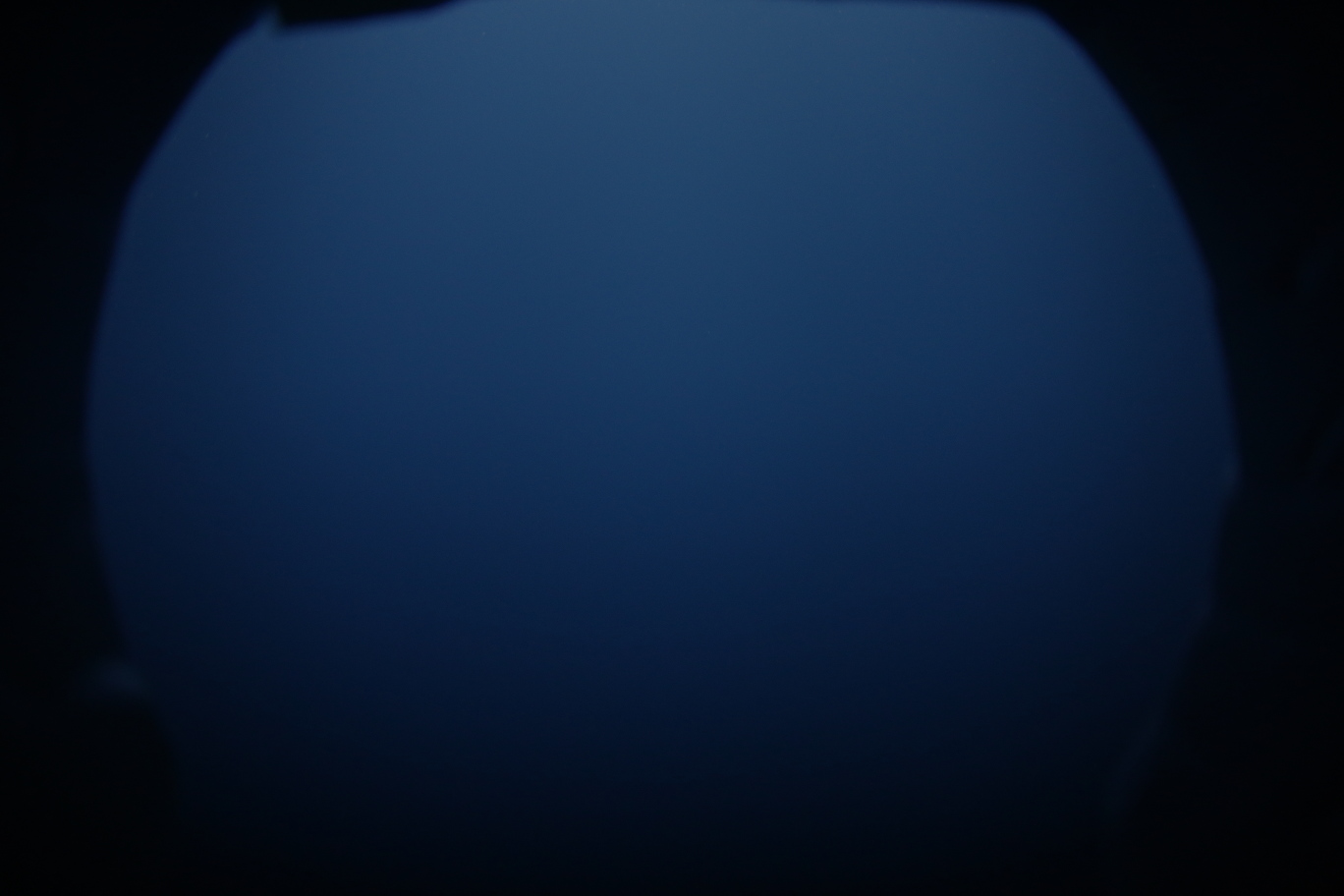} 
	\caption{A water column image $I_c(x,P_\mathrm{high\; altitude})$ taken by the AUV very high above the seafloor. It only shows scattering effects.}
	\label{fig:rawandscatter}
\end{figure}

\begin{figure}[h]
	\centering
	\includegraphics[width=0.19\columnwidth]{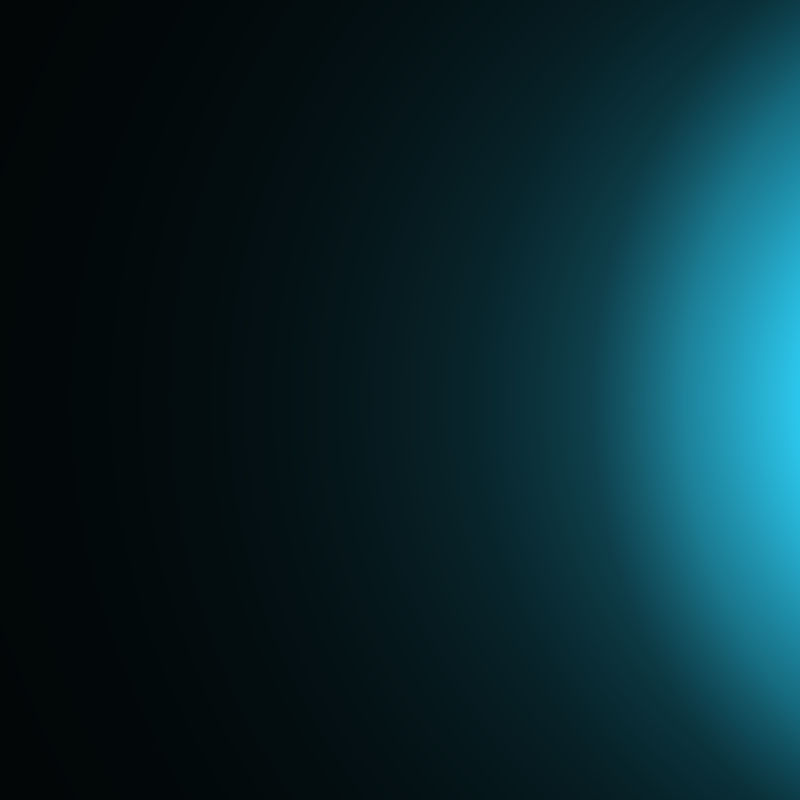} 
	\includegraphics[width=0.19\columnwidth]{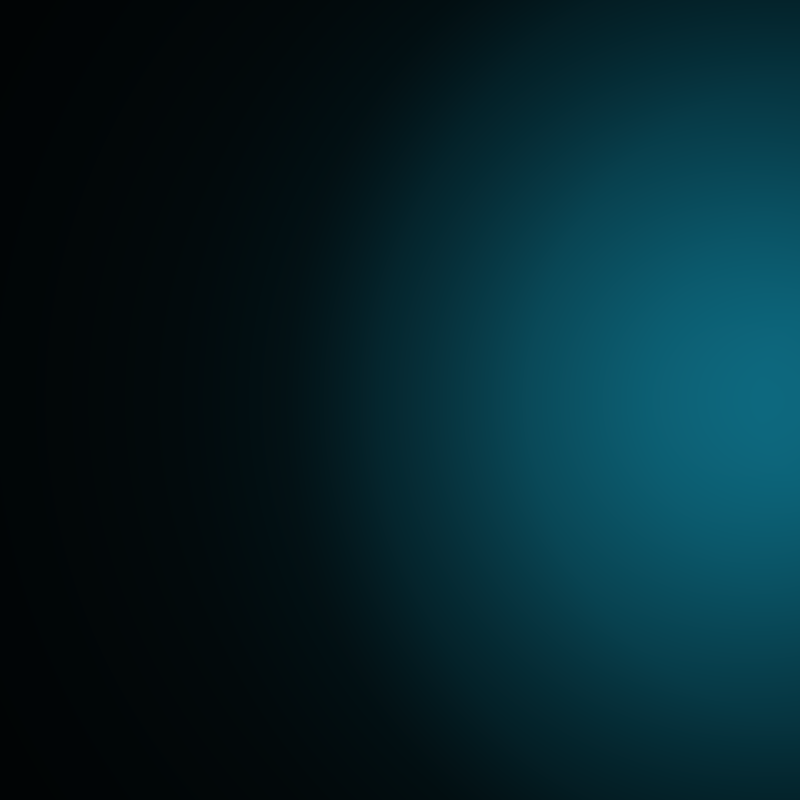} 
	\includegraphics[width=0.19\columnwidth]{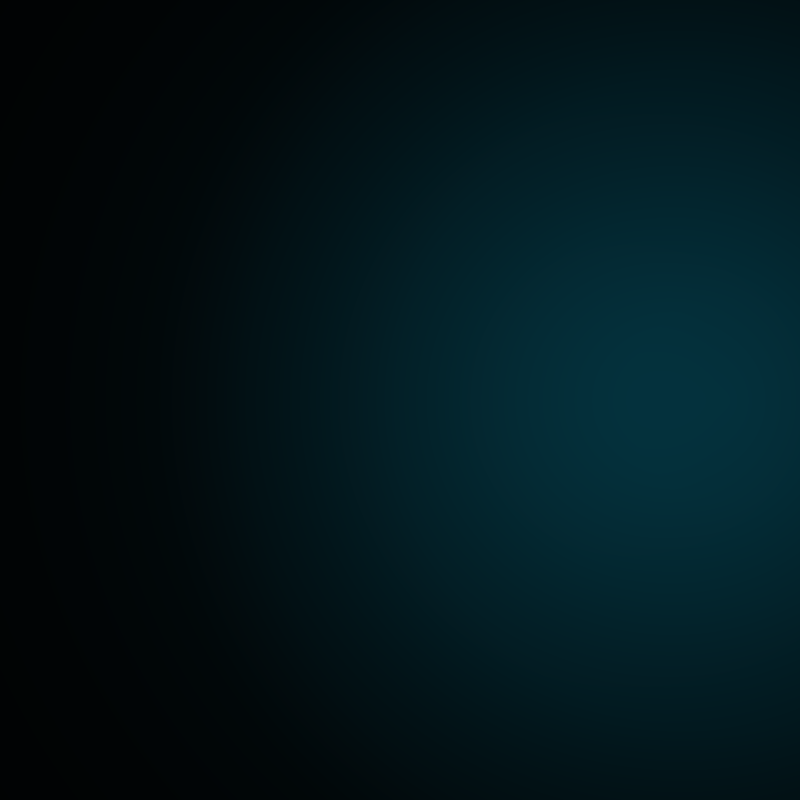} 
	\includegraphics[width=0.19\columnwidth]{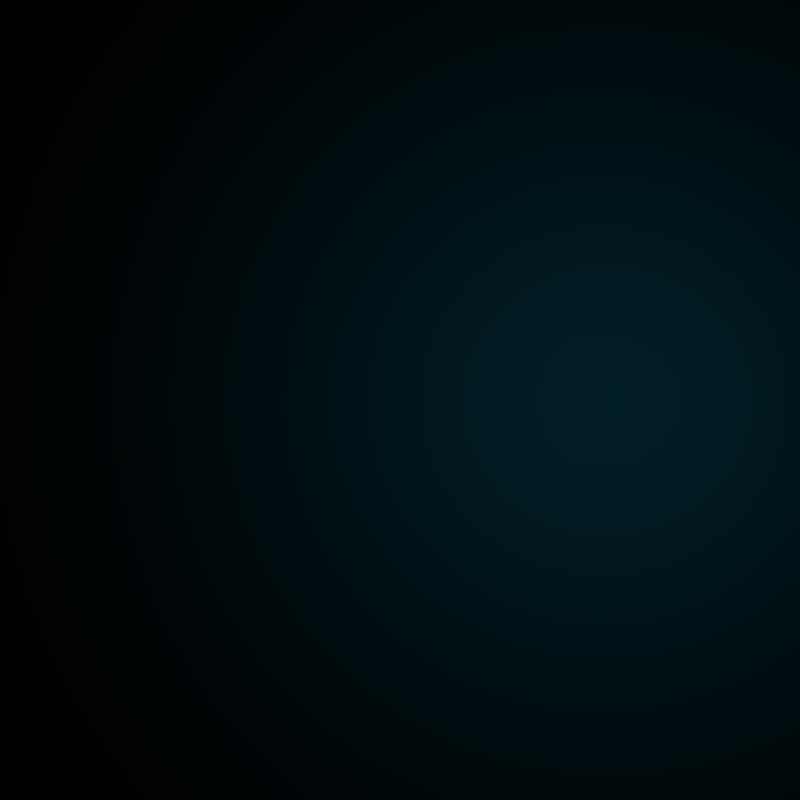} 
	\includegraphics[width=0.19\columnwidth]{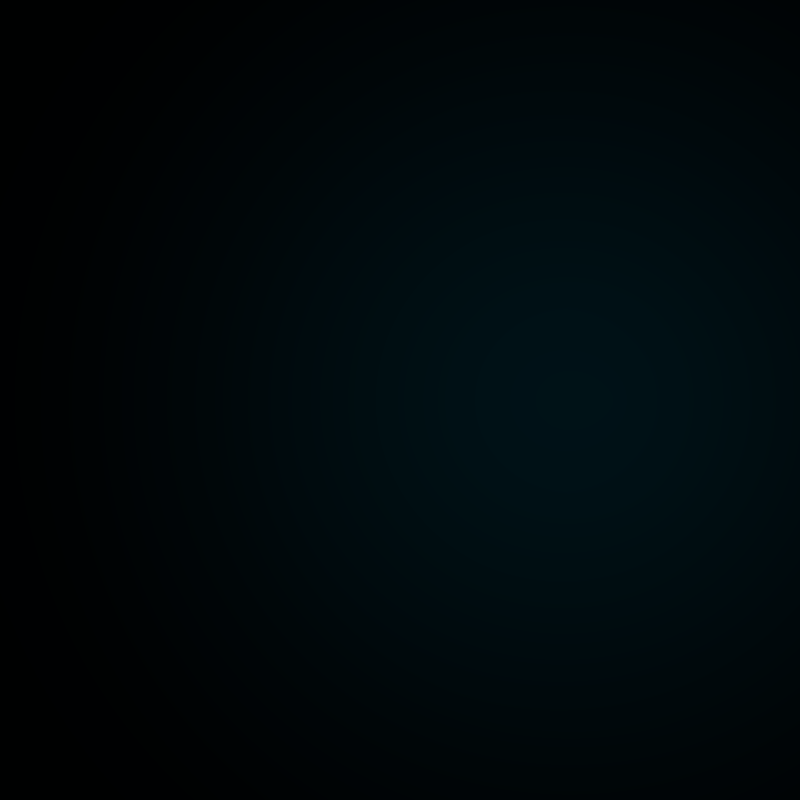} 
	\caption{For a light source 2m to the right of a camera (Gaussian light cone), looking into the same direction, the scatter in Jerlov type II water is shown, as it reaches the camera from 1m (left image) to 5m (right image). It can be seen that relatively little light is scattered into the camera sensor from a distance of 5m. This trend continues with further distances, such that scattered light will be insignificant after a certain distance \cite{song2021deepsea}. This illustration considers single scattering only.}
	\label{fig:scatteringrendered}
\end{figure}

\begin{figure}[h]
	\centering
	\includegraphics[width=0.48\columnwidth]{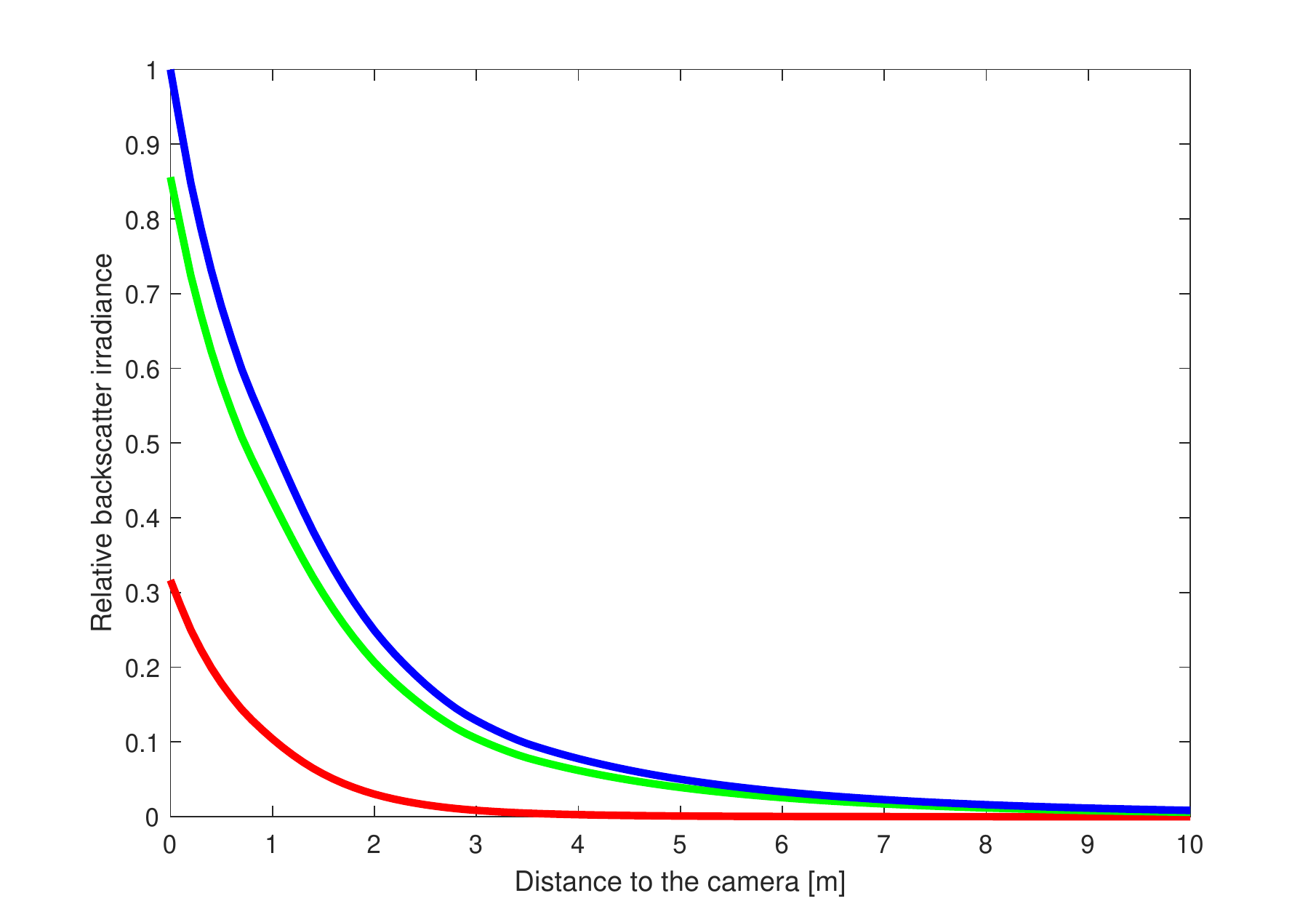}
	\includegraphics[width=0.48\columnwidth]{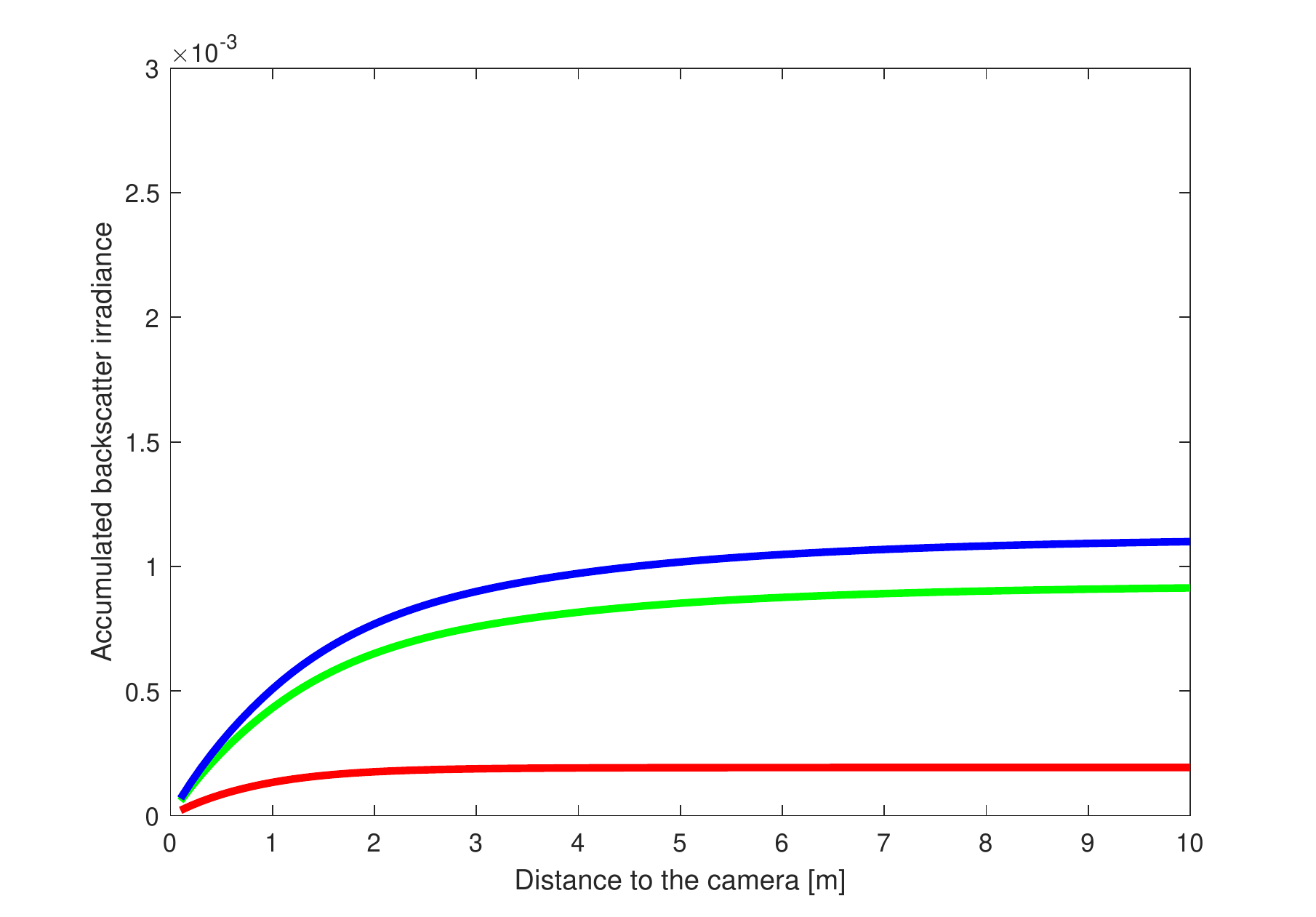}
	\includegraphics[width=0.48\columnwidth]{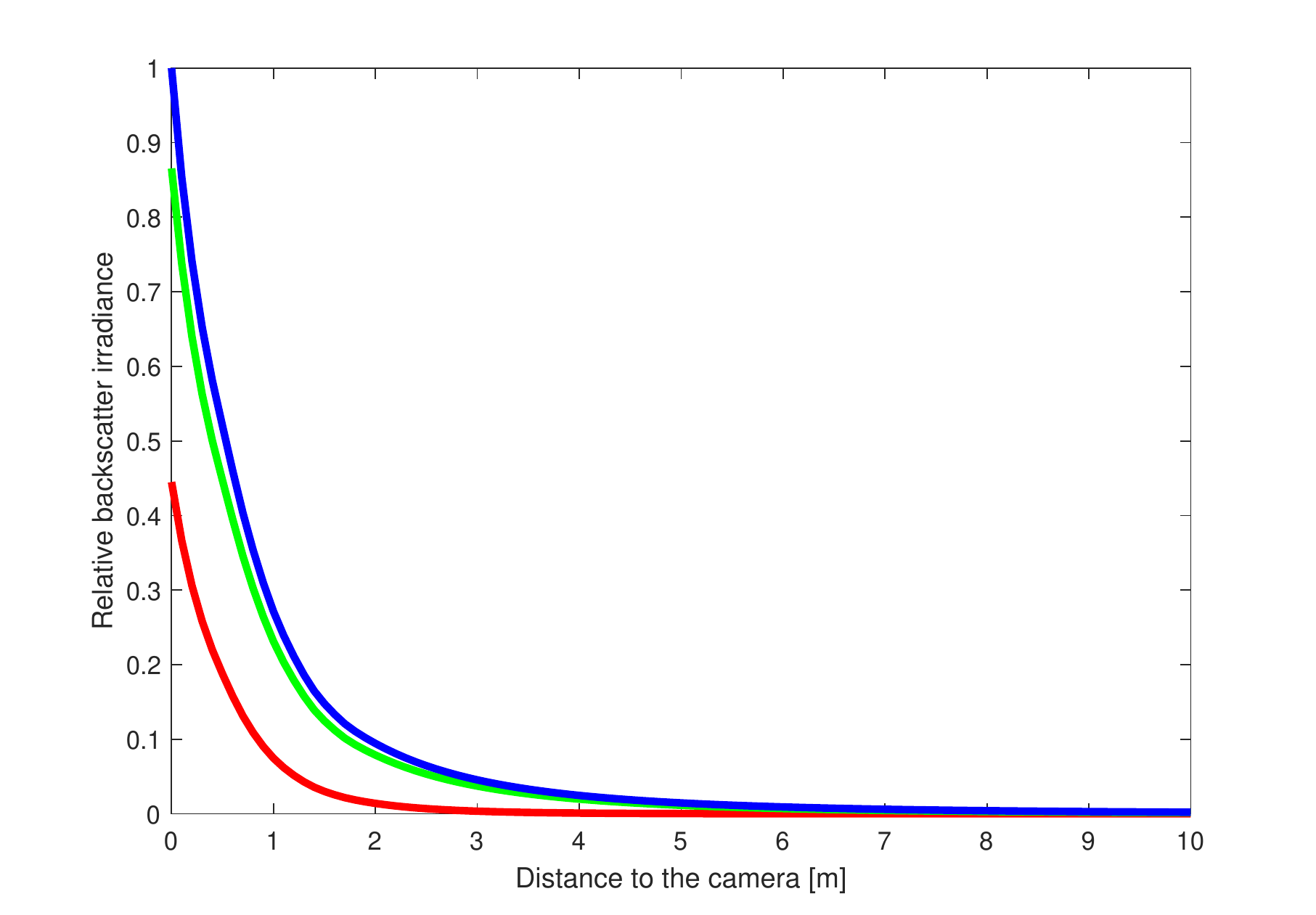} 
	\includegraphics[width=0.48\columnwidth]{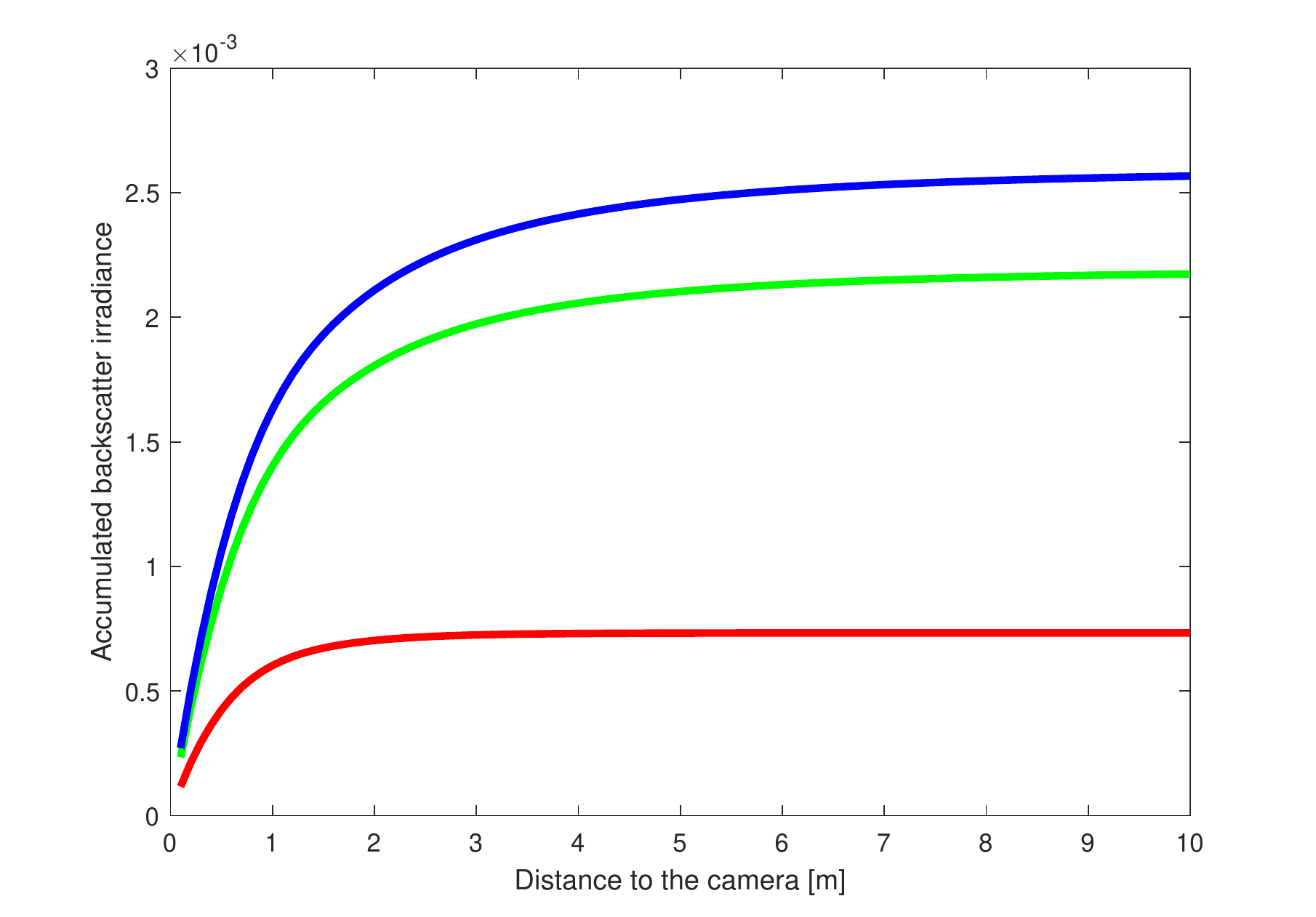}
	\caption{Ideal point light source positioned at 2m (top row) and 1m (bottom row) lateral distance from a camera in Jerlov water type II. Left: Irradiance observed by a camera pixel from scattering at given distances (horizontal axis). Right: Cumulative curve showing how much scattered light is collected along the viewing ray up to a certain distance. It can be seen that the curve saturates soon, i.e. most of the scattered light originates from the first few meters.}
	\label{fig:scatteringanalysis}
\end{figure}

For an image taken at very high altitude (see fig. \ref{fig:rawandscatter}) $d_l$ and $d_c$ will be so large that $F(x,P_{\mathrm{high\; altitude}}) \approx 0$ will hold and we will see only backscatter:
\begin{equation}
I_c(x,P_\mathrm{high\; altitude}) = I_s(x,P_\mathrm{high\;  altitude}),
\label{eq:backscatter}
\end{equation}
This backscatter actually depends only on the relative pose of the light source with respect to the camera (but not to the ground) and the angular characteristics of the light source. To distinguish the deep sea setting from the natural sunlight setting, in fig. \ref{fig:scatteringrendered} we applied the simulator from \cite{song2021deepsea} to simulate the scattering that happens in the deep sea (or at night) at different distances to the camera when a light cone originates from a position 2m to the right of the camera in the open water. This is a simple motivational example that only considers single scattering between light source and camera and the volume scattering function is chosen from Petzold's measurement \cite{petzold1972volume} (clear water), but attenuates all light exponentially with the distance travelled in the water.
It can be seen that most of the visible scattering happens close to the camera, because also the scattered light is attenuated and only little intensity is observed from far away scattered light.
There is only relatively little light scattered at 5m distance that reaches the camera. 
In fig. \ref{fig:scatteringanalysis} we plot the backscatter received by a hypothetical camera along a single viewing ray where the light is at 2m and 1m distance sideways to the camera.
For this simulation we use Jerlov water type II and volume scattering according to Petzold \cite{petzold1972volume} (offshore southern California). 
However, in our experience, this scenario holds also when operating with artificial illumination in murky water, simply with all distances reduced: The camera has to go closer to the seafloor in order to see it, and for practical reasons (keep homogenoeus seafloor illumination, avoid drastic shadows, robot maneuverability close to the ground) we then use light sources that are closer to the camera (deep sea also requires more massive, larger vehicles), resulting in a similar relative geometry between seafloor, camera and light.

Consequently, we now make the assumption that the majority of the scattering originates from the first few meters in front of the camera and that during the seafloor mapping we will always fly "high enough" to see most of the scattering, i.e. $I_s(x,P)\approx I_s(x,P_\mathrm{high\; altitude})$. Substituting this back into the equation, we obtain:
\begin{equation}
I_c(x,P) \approx A_s(x,P) \cdot F(x,P) + I_c(x,P_\mathrm{high\; altitude}),
\label{eq:simplified_replaced}
\end{equation}
i.e.\ the intensity observed in the camera consists of the seafloor albedo multiplied by a factor image $F$ that depends on the lighting configuration, the relative pose of the vehicle with respect to the seafloor and the water attenuation, plus a fixed scatter image.
For optical deep sea mapping missions in smooth terrain, AUVs usually follow a "fixed altitude" mission. In this case the pose $P$ is constant over time and the factor image $F$ just depends on the pixel position but does not vary over time. But even if the AUV varies the altitude, the motion change of heavy diving robots is usually so slow that $F$ can be considered almost constant over short periods of time.
In order to infer the seafloor color $A_s$ from an image $I_c$, we will now perform robust estimates of the "factor image" $F$ and the summand image $I_c(x,P_\mathrm{high\; altitude})$.

\subsection{Estimation}
\subsubsection{Additive Term}
Before reaching the working altitude above the seafloor the robot should already capture a certain number $b$ of images that only show the water column, revealing information about the scattering. In practice, these images will also contain bright floating particles or dark parts very close to the camera which are not inside the light cone. These measurements have to be considered as outliers and consequently a robust estimator (cf. to \cite{robuststatistics}) is required to obtain $I_s$ from multiple measurements at each pixel position. We focus on the case where much less than half of the image pixels show floating particles, and where the particle positions can be considered random (i.e. they are not staying at the same position over time). This means that at each pixel position more than 50\% of the time we can observe $I_s$. Since the median has a breakdown point \cite{robuststatistics} of 50\%, we will perform a temporal median (across the $b$ measurements) at each pixel position $x$ to infer the ideal pure scatter image $I_c(x,P_\mathrm{high\; altitude})$. 

\subsubsection{Multiplicative Term}
We now suggest to estimate the factor image $F$ in a similar way.
For illustration, imagine first that an ``all-seafloor image'' $I_{\mathrm{all-seafloor}}$ would be given that shows only homogeneous sediment of known sediment color $A_{\mathrm{sediment}}$. In such an ideal image of a homogeneous seafloor, we would typically still see the illumination pattern: The factor image $F$ would cause different observed pixel values depending on the pixel position, and each pixel in $F$ represents the factor for the light ray that belongs to the corresponding viewing ray. We denote the pose when this all-seafloor image was seen by $P_a$, such that we can write also this ideal all seafloor image using multiplicative and additive terms:
\begin{equation}
\label{eq:calibration}
I_{\mathrm{all-seafloor}}(x,P_a) = A_{\mathrm{sediment}} \cdot F(x,P_a) + I_c(x,P_\mathrm{high\; altitude})
\end{equation}
This can be solved linearly for $F$ at each pixel position $x_I$, since all other components are given. Afterwards, both the additive and the multiplicative lighting effects are known.

Now, whenever the AUV has exactly this relative pose $P_a$ to the seafloor, we can compute the seafloor albedo at a particular image position $x_I$ by the simple division
\begin{equation}
\label{eq:normalization}
A_s(x) = (I_c(x, P_a)-I_c(x,P_\mathrm{high\; altitude})) / F(x, P_a)
\end{equation}
$A_s(x)$ is our restored image of the seafloor.

It can be seen that the seafloor color $A_{\mathrm{sediment}}$ used in equation \ref{eq:calibration} plays the role of a white balance reference in normal photographs. When setting it to grey although the actual seafloor color is brown, all other colors change accordingly, but in a consistent linear fashion. $F$ is still correct up to a global scale factor, i.e.\ we can later also correct all images up to one single global scale factor (per color channel resp.\ wavelength) if desired. For mapping and reconstruction this means that without prior knowledge the seafloor color can be chosen as grey and all images will be enhanced in a consistent way (allowing matching, SLAM and stereo reconstruction). This is similar to mapping on land with a camera that uses a fixed but unknown white balance.
\begin{figure}[ht]
	\centering
	\includegraphics[width=0.485\columnwidth]{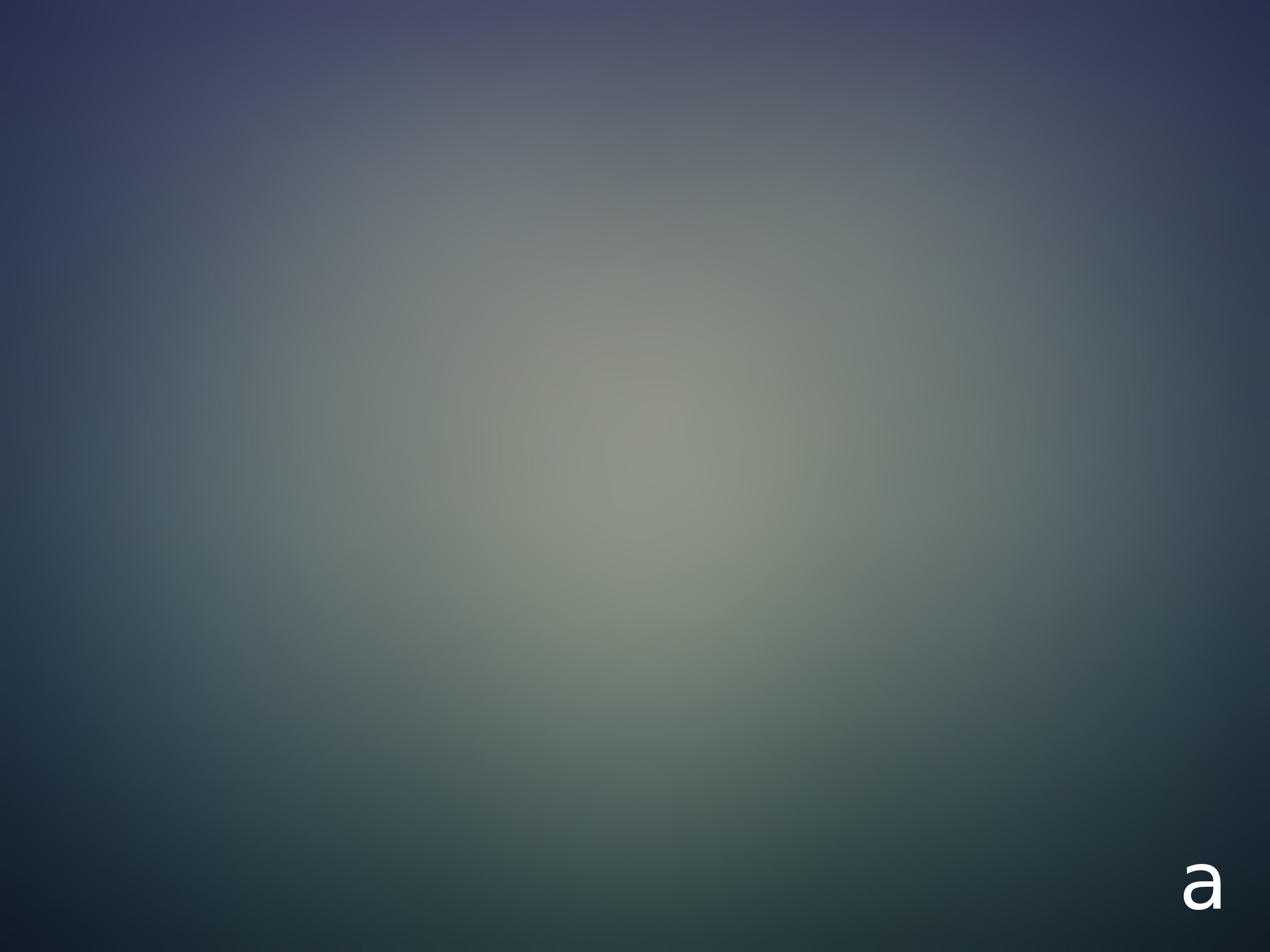} 
	\includegraphics[width=0.485\columnwidth]{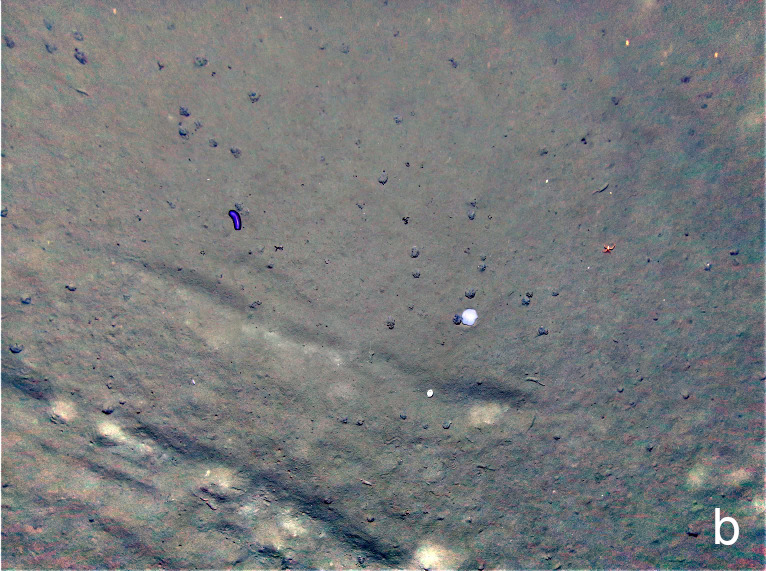}
	\includegraphics[width=0.32\columnwidth]{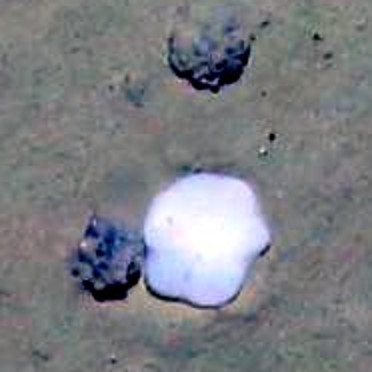}
	\includegraphics[width=0.32\columnwidth]{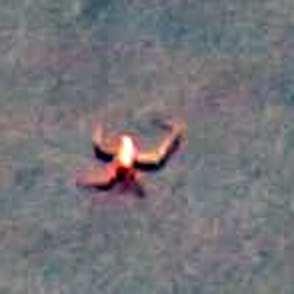}
	\includegraphics[width=0.32\columnwidth]{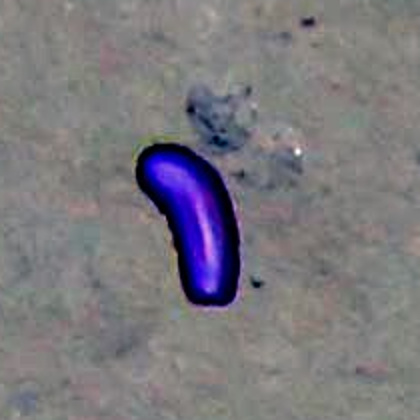}
	\caption{A virtual all-seafloor image, where all objects are robustly suppressed (a) encodes the lighting setting of the current image. When approximately removing the illumination effects using the all-seafloor image, we obtain the normalized image (b).
		Three regions of the normalized image have been zoomed in to provide an idea of the resolution (bottom). See fig. \ref{fig:rawimage} for the raw image.}
	\label{fig:normalization}
\end{figure}

Since usually a perfect all-seafloor image as needed by equation \ref{eq:calibration} is not available, we will now consider estimating it from survey data.
For instance, in case multiple all-seafloor images exist, each perturbed by Gaussian noise, it is suggested to average them for estimation of $F$.
These images can even be captured at different locations, as long as the relative pose between the AUV and the seafloor ground plane stays the same (same altitude and pitch and roll).
In the common case the seafloor is not of uniform color, but ``contaminated'', e.g.\ by stones, fauna or other objects that cannot be considered Gaussian noise, a robust estimate of the all-seafloor image is suggested.
Same as for the scatter image, for scenarios where the images shows significantly more than 50\% uncontaminated seafloor with a uniform spatial distribution of objects occuring, we suggest using the temporal median as a robust estimator at each pixel position, i.e. we compute the median intensity at each pioxel position over many images.
The majority will contain seafloor at this coordinate, and only in a few images this pixel will display a rock or other object. In the next section we will turn to the question over how many images we have to compute the median at least.

\subsubsection{Determining the number of samples}
Let $n$ be the number of images where we inspect a certain image position $x_I$ and let $c \in ]0;0.5[$ be the general contamination rate of the images.
The probability $P$ of obtaining $u$ uncontaminated seafloor samples from $n$ images is then described by a binomial distribution $\mathcal{B}(c, u, n)$. 
With increasing number of images $n$, the probability $P_{\mathrm{half}}$ that at least half the samples being contaminated becomes smaller and smaller (as $c<0.5$):
\begin{equation}
P_{\mathrm{half}} = \sum_{u=n/2}^{n} \mathcal{B}(c, u, n) 
\label{eq:mediansize}
\end{equation}
Consequently, the number $n$ of contaminated seafloor images to be used in pixelwise median estimation depends on the contamination rate $c$, and can be chosen such that $P_{\mathrm{half}}$ from eq. (\ref{eq:mediansize}) becomes almost zero.
For instance, if $c=20\%$ of the image is contaminated and using a median on 7 images, only in $P_\mathrm{half} \approx 3\%$ of the cases more than 3 contaminated samples are expected, which would invalidate the median estimate at this position (since the median has a breakdown point of 50\%). Still, 3\% of a 10 megapixel image means 300.000 pixels, for which median estimation does not work. For very special, high-frequent illumination patterns, the number of images must be increased in this case. However, typical illumination patterns vary smoothly, and so $F$ can be expected to be almost constant in a small neighborhood around some position $x_I$. Therefore, the number of samples can also be increased by including spatial neighbors of the pixel under consideration and robust spatial averaging {\em within the image} (i.e. a spatial median) can be used to increase the number of uncontaminated seafloor samples available. 
% In case of smooth spatial illumination (no high-frequent changes of illumination pattern), it is possible to perform the estimation of the pattern on a smaller resolution to speed up computation. Consequently, we propose to use the spatial Median as a robust smoother prior to downsampling to create a Median-pyramid of the image.

\subsubsection{Varying Poses and Terrain}
Unfortunately, because of terrain variations, and because of altitude and pose variations of the vehicle, the relative pose between the camera and the seafloor typically varies during a several hours mission. Therefore it is not recommended to compute only one all-seafloor image for an entire image sequence, but to compute an individual all-seafloor image in a sliding window fashion for each image, as for short amounts of time, the pose of the vehicle with respect to the seafloor is typically stable.

To summarize, for scenarios with $c \ll 50\%$ seafloor contamination, we suggest to compute a sliding window temporal median for the $n$ images before, at and after the current image under investigation. Then, each resulting all-seafloor image is used in eq. (\ref{eq:normalization}) for normalization of the respective original image after subtracting the additive scatter commponent.

% median: runtime on pyramid for cpu implmenmtation, mehrheitswahlrecht usa, formel abschätzung für kernel size n, wenn auf 70prozent gezielt wird, wie darf die lokale häufigkeit sich verändern bis ein artefakt ins bild kommt

\section{Results}

\begin{figure}
	\centering
	\includegraphics[width=0.32\columnwidth]{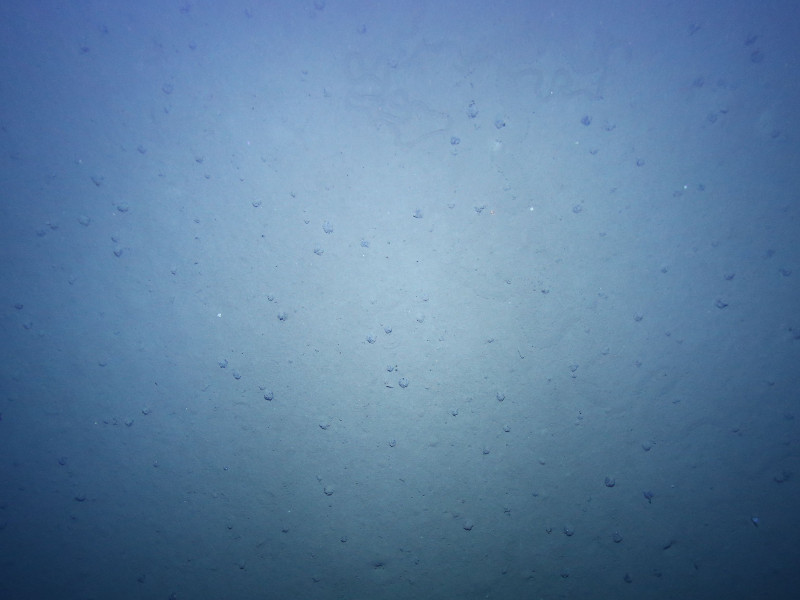} 
	\includegraphics[width=0.32\columnwidth]{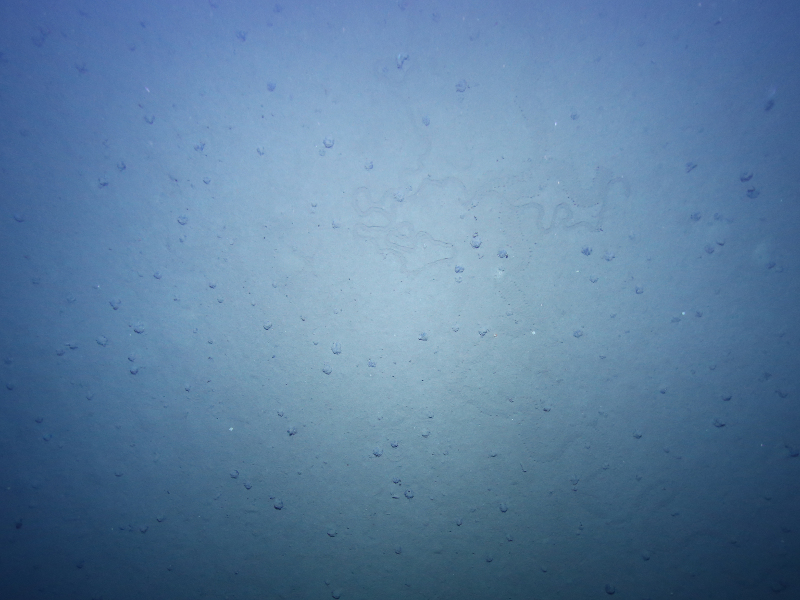} 
	\includegraphics[width=0.32\columnwidth]{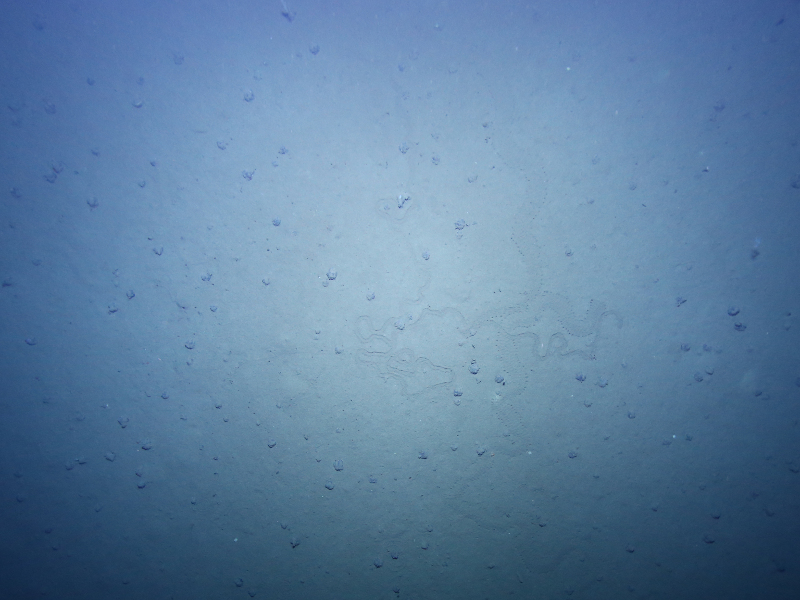} \\
	\includegraphics[width=0.485\columnwidth]{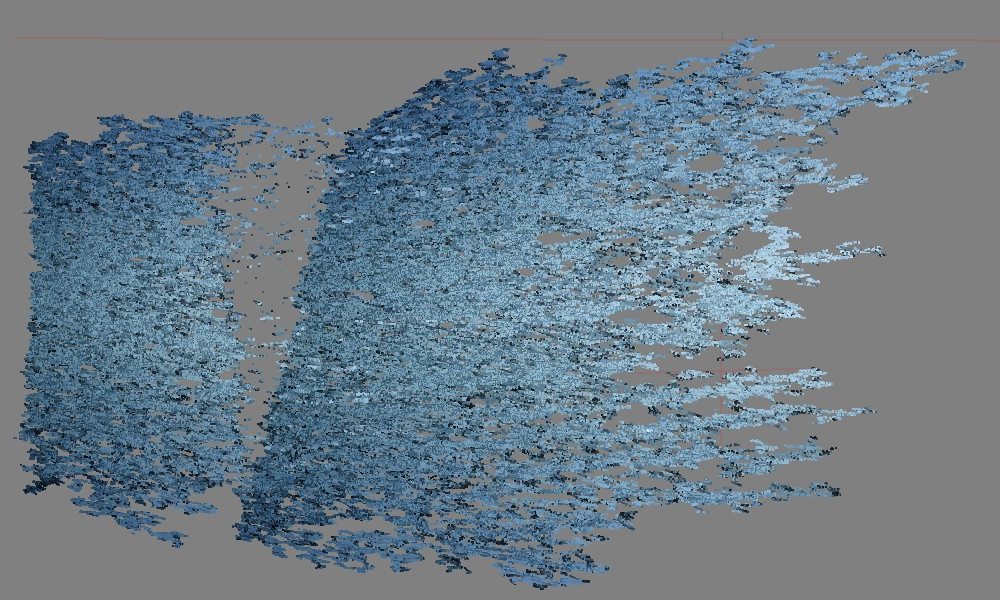} 
	\includegraphics[width=0.485\columnwidth]{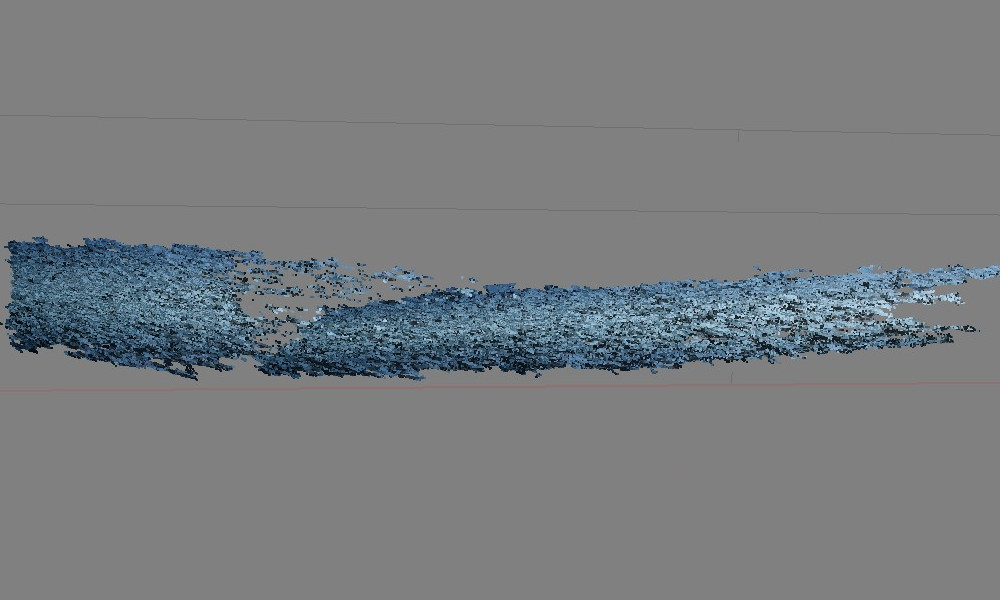} \\
	\includegraphics[width=0.485\columnwidth]{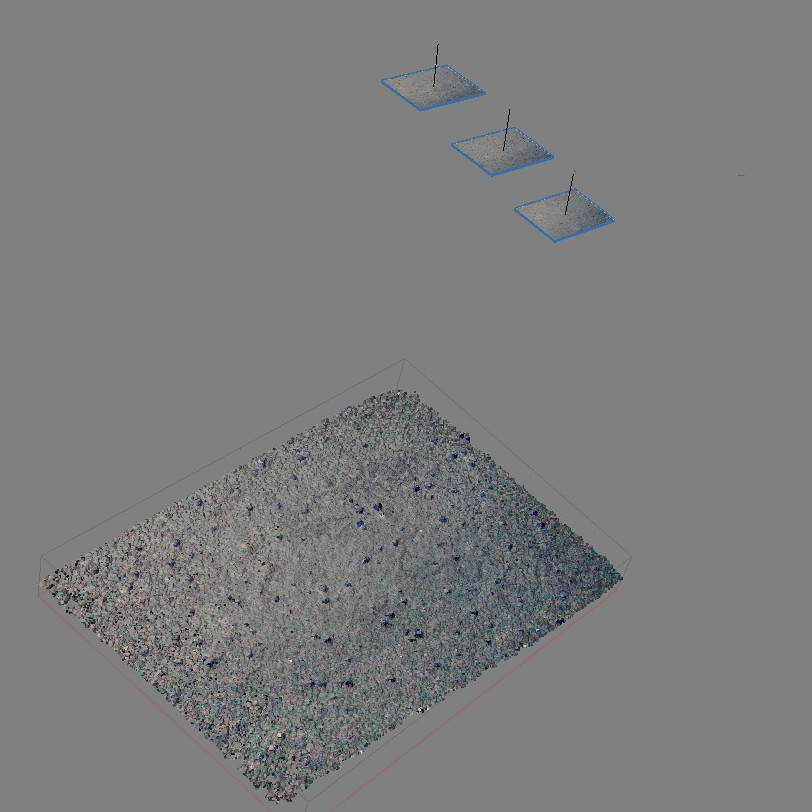} 
	\includegraphics[width=0.485\columnwidth]{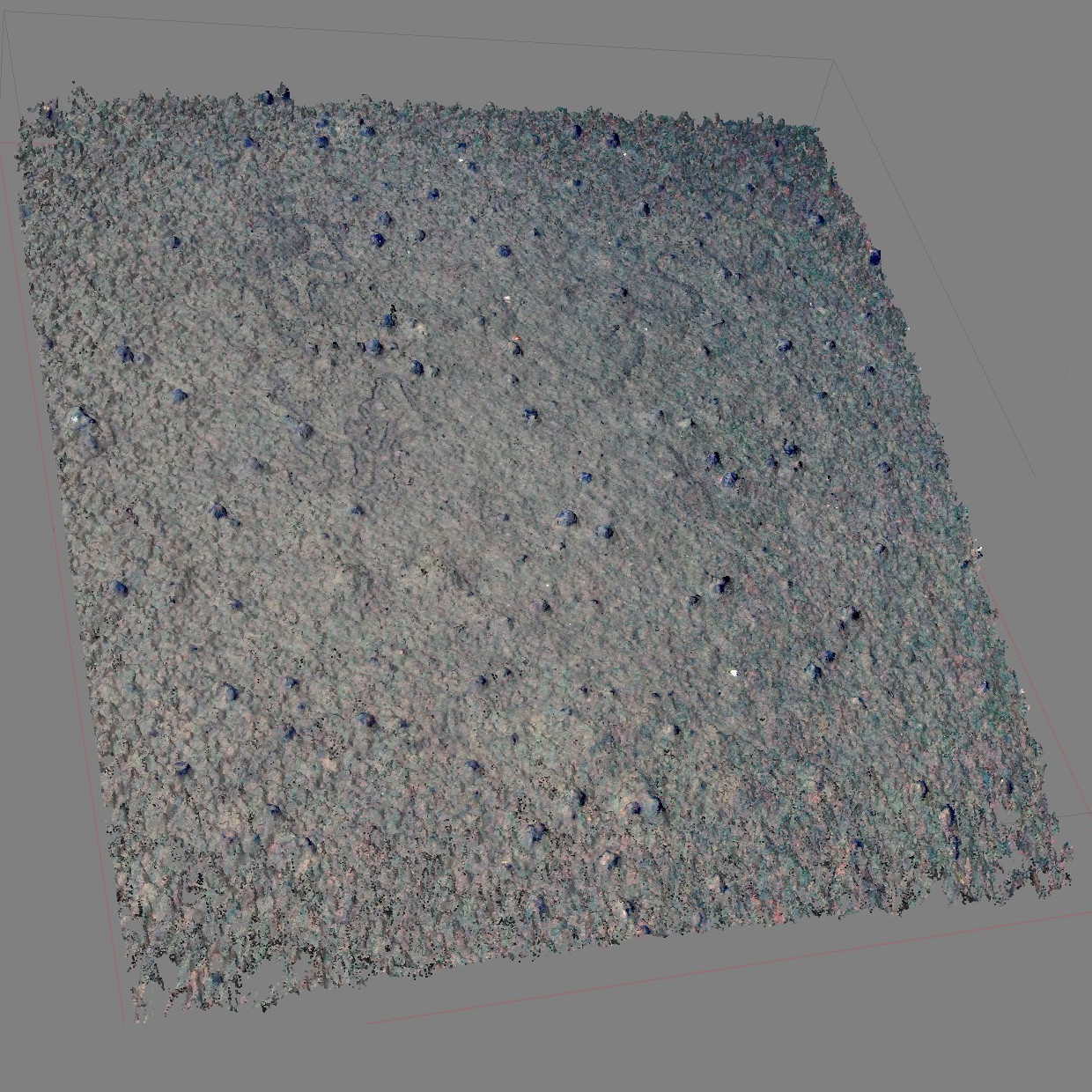} \\
	\includegraphics[width=0.32\columnwidth]{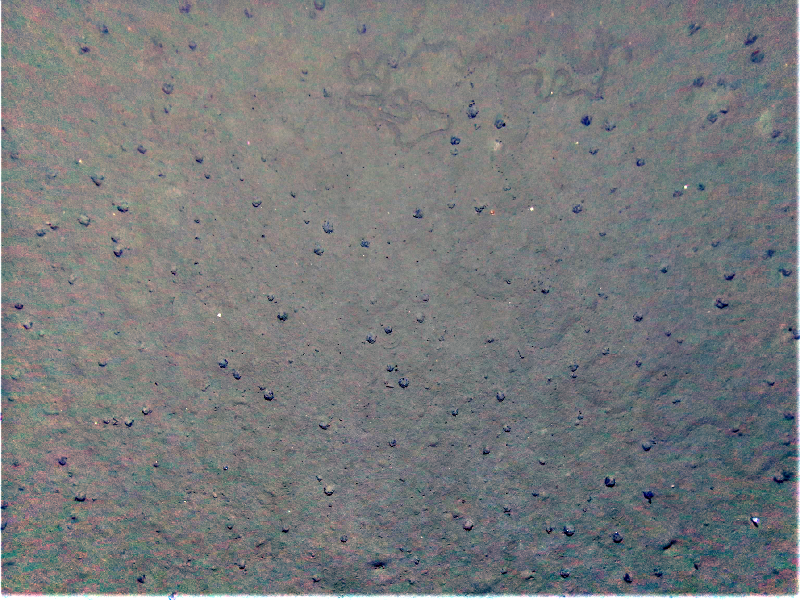} 
	\includegraphics[width=0.32\columnwidth]{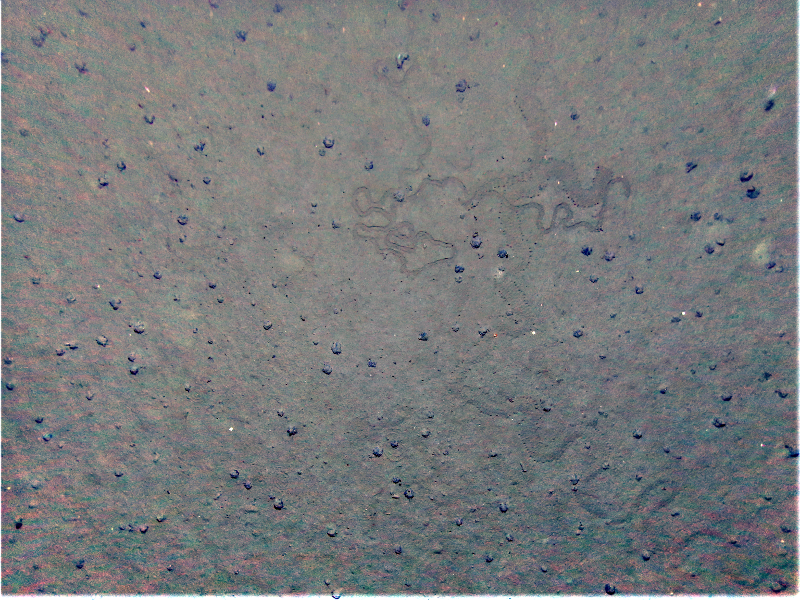} 
	\includegraphics[width=0.32\columnwidth]{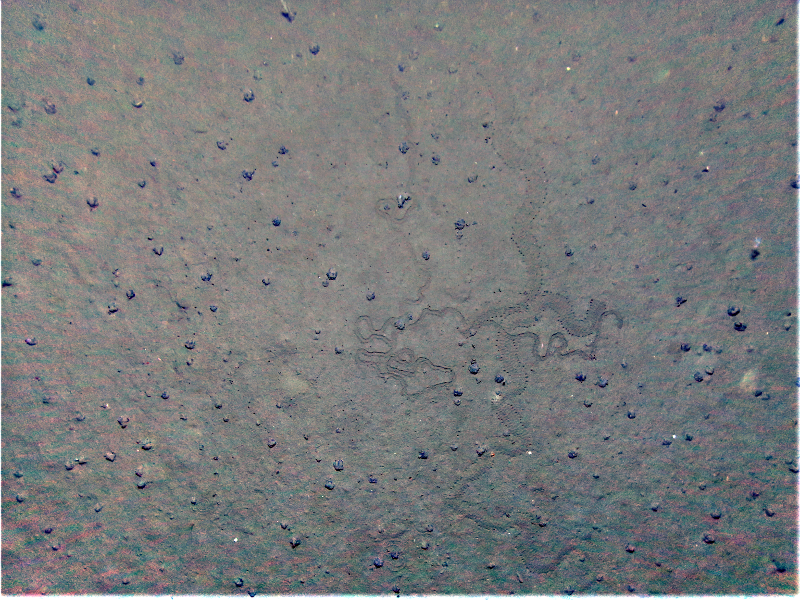} \\
	
	\caption{Top row: Failed 3D reconstruction from three unnormalized blueish photos. Bottom row: Correct 3D reconstruction from the normalized versions of the same images.}
	\label{fig:3d}
\end{figure}

\subsubsection*{Implementation}
The goal of this work is an approach that can efficiently handle ten-thousands of photographs to enable large scale deep ocean mapping. The prototypical, unoptimized reference implementation on a single 3.7GHz Xeon CPU requires almost one minute of computation time for each 12 Megapixel photo. However, most operations are suitable for parallel scheduling and thus the algorithm was implemented in CUDA and parallelized on the pixel level. All images needed for the temporal median are uploaded to the GPU at the same time and organized as a ring buffer. Once the median on the central image is completed, the "oldest" image is replaced by a new image from the stream.

To reduce the number of temporal images required for reliable robust estimation, and since the illumination patterns are usually very smooth, by default we downsample the undistorted 12MP input images by a factor of 8 (after a spatial median) to perform the temporal median on 7 images (and upsample $F$ afterwards). The overall implementation provides estimates of $F$ and produces 12Megapixel color-corrected images at 2Hz, which is twice as fast as we record the photos during a mission.
%\begin{figure}[t]
%	\centering
%	\includegraphics[height=18cm]{{track007/mosaic-unnormalized}.png} 
%	\includegraphics[height=18cm]{{track007/mosaic-normalized}.png}
%	\caption{Deep sea mosaic stitched from raw photos (left) and enhanced (right).}
%	\label{fig:mosaicunnormalized}
%	\label{fig:mosaicnormalized}
%\end{figure}

%(2) FusionEnhance: https://ieeexplore.ieee.org/document/6247661
%\cite{ancuti12enhancing}
%(3) MultiExposureFusion: %https://www.sciencedirect.com/science/article/abs/pii/S0165168418301063
%\cite{galdran18dehazingmulti-exposurefusion}
%(4) OptimizedContrastEnhance: %https://www.sciencedirect.com/science/article/pii/S1047320313000242
%(5) RemoveBackScatter: master thesis
%\cite{zhang2016removing}

%./different_uw_img_enhancement/199papersample/mosaic_20200731_130951/mosaic.png-cropped.jpg
%./different_uw_img_enhancement/199papersample_OptimizedContrastEnhance/mosaic_20200731_130138/mosaic.png-cropped.jpg
%./different_uw_img_enhancement/199papersample_FusionEnhance/mosaic_20200731_125538/mosaic.png-cropped.jpg
%./different_uw_img_enhancement/199papersample_RemoveBackScatter/mosaic_20200731_130457/mosaic.png-cropped.jpg
%./different_uw_img_enhancement/199papersample_MultiExposureFusion/mosaic_20200731_125844/mosaic.png-cropped.jpg
%./199papersample/scattersubtracted/mosaic_20200731_125049/mosaic.png-cropped.jpg
%./199papersample/brown/mosaic_20200731_124318/mosaic.png-cropped.jpg
%./199papersample/grey/mosaic_20200731_124715/mosaic.png-cropped.jpg
%./mosaic_20200731_110040/mosaic.png-cropped.jpg

\subsection{3D Reconstruction}
\begin{figure}
	\centering
	\includegraphics[width=0.48\columnwidth]{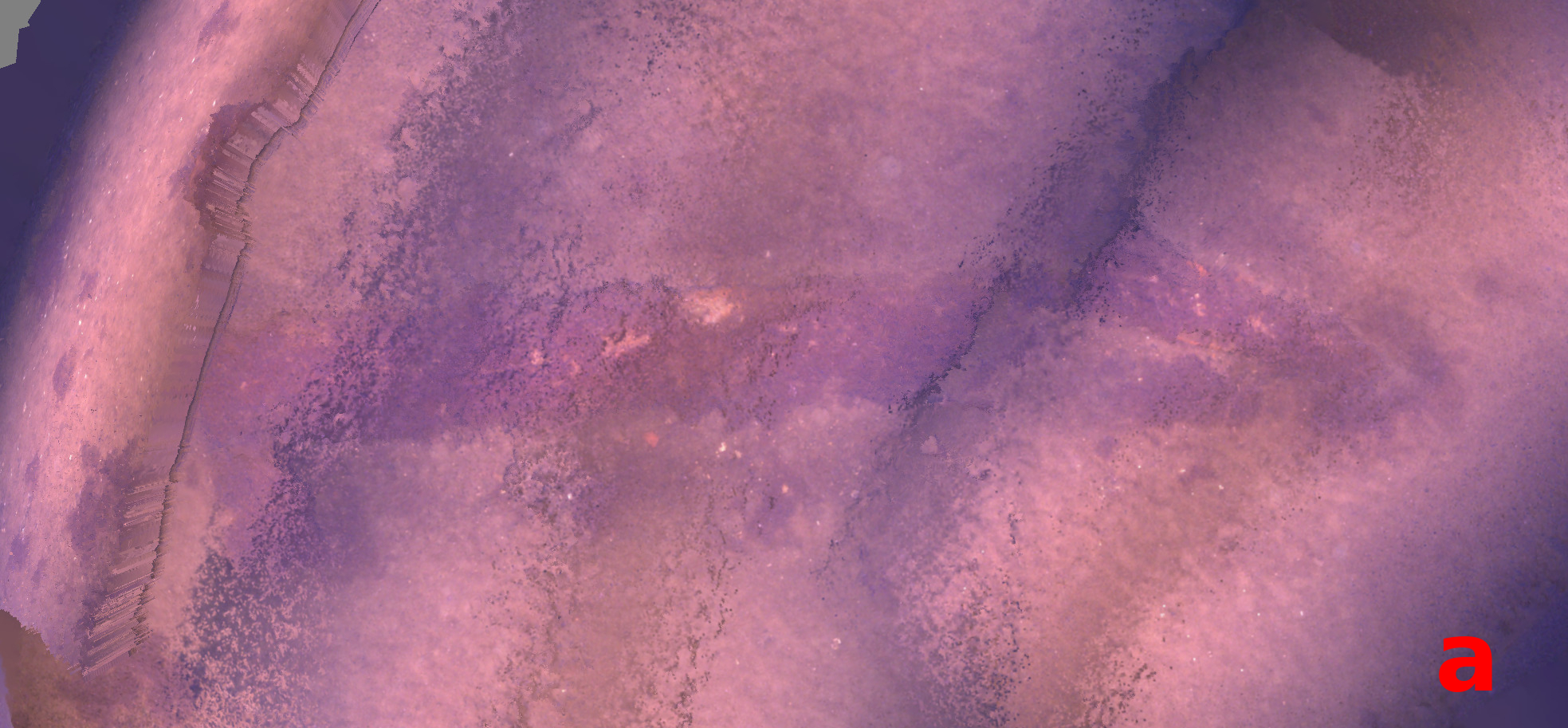} 
	\vspace{0.5mm} 
	\includegraphics[width=0.48\columnwidth]{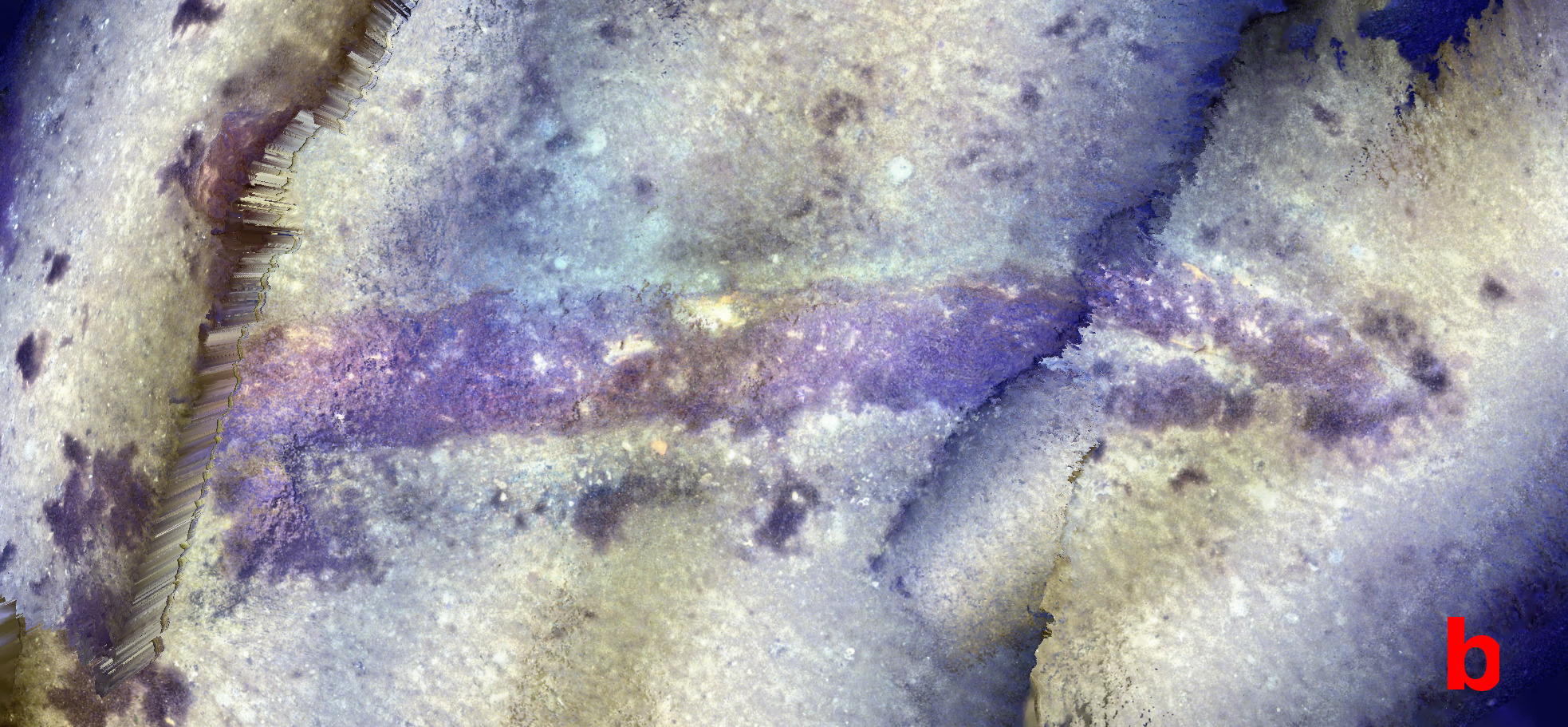} 
	\vspace{0.5mm}
	\includegraphics[width=0.48\columnwidth]{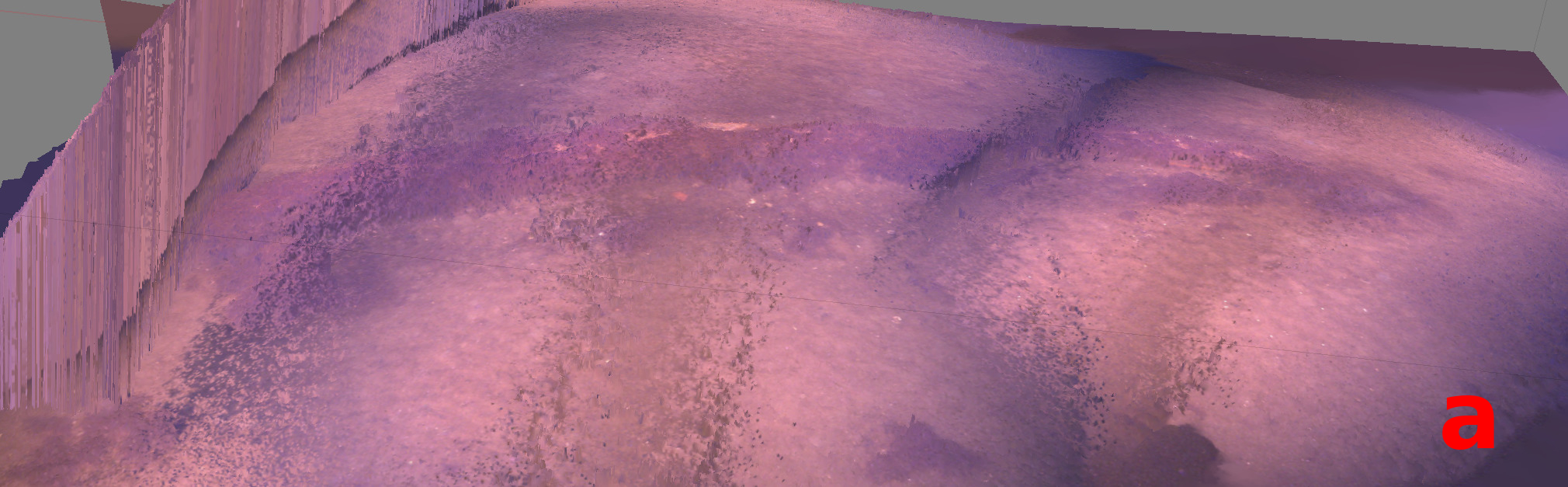}
	\vspace{0.5mm}
	\includegraphics[width=0.48\columnwidth]{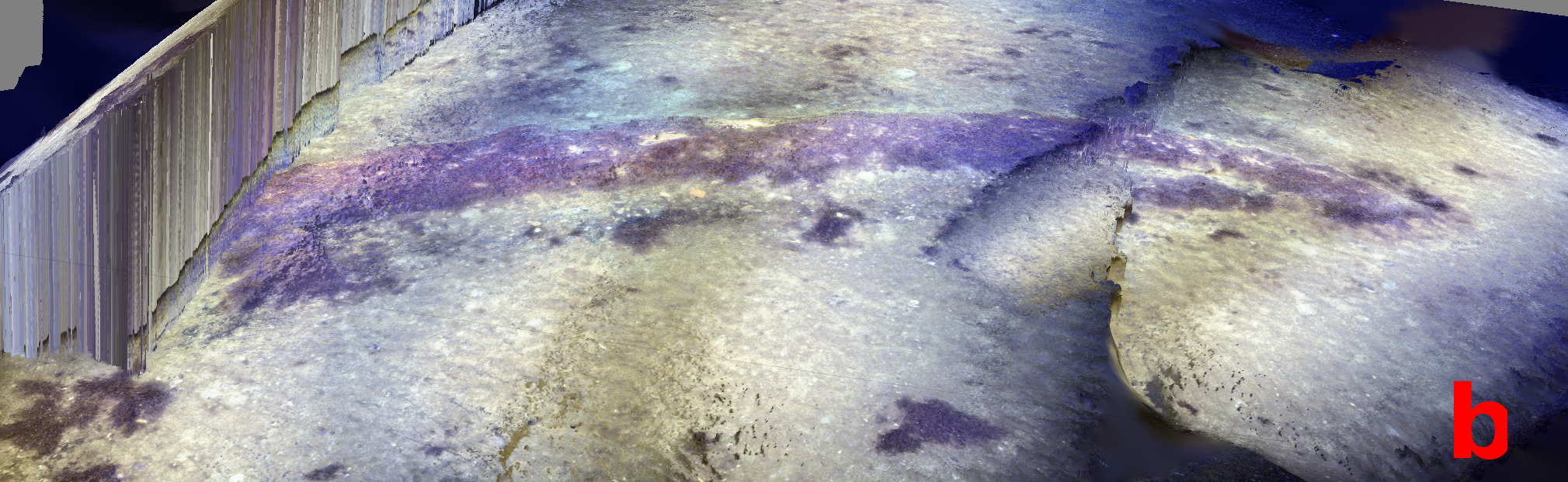} 
	\vspace{0.5mm} 
	\includegraphics[width=0.48\columnwidth]{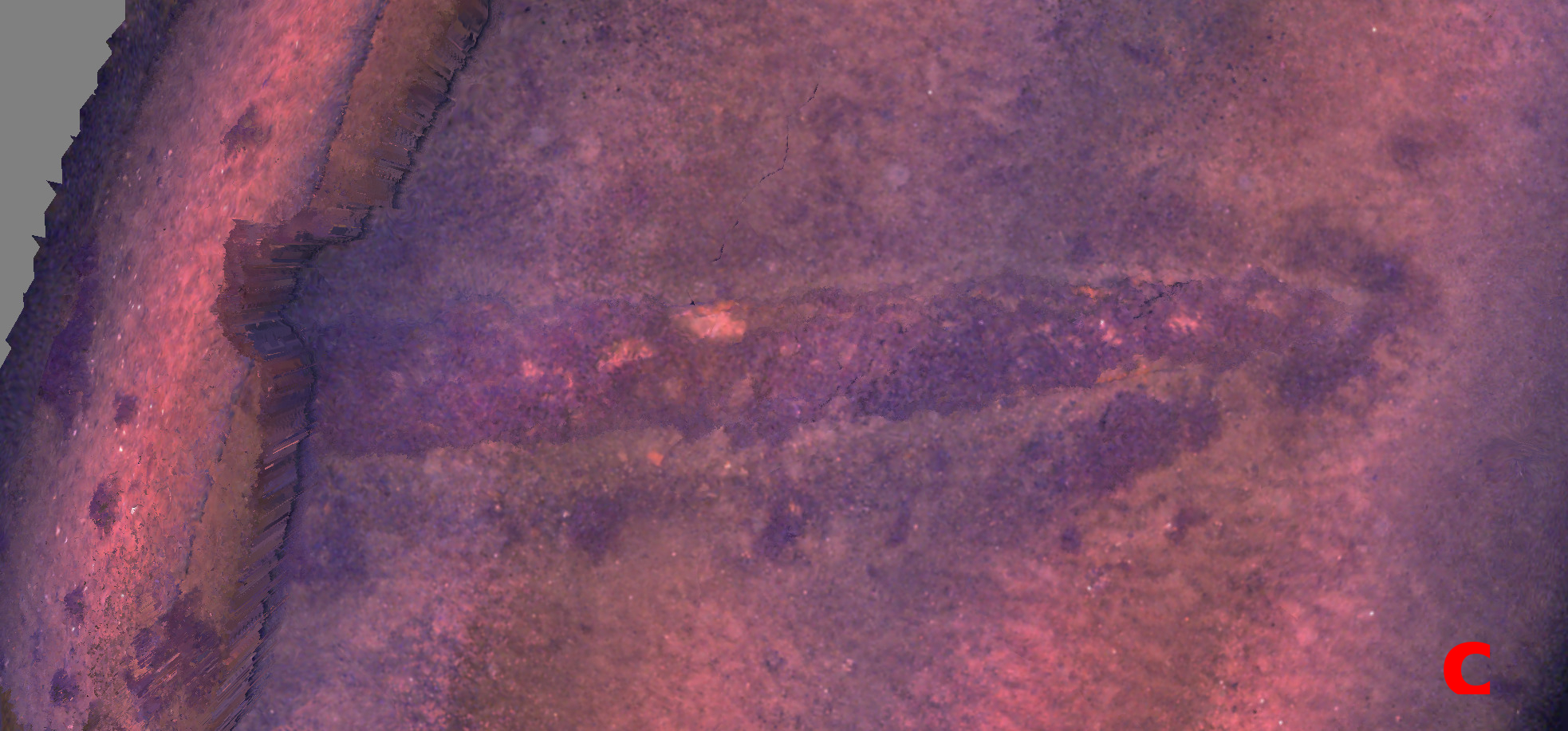}
	\vspace{0.5mm} 
	\includegraphics[width=0.48\columnwidth]{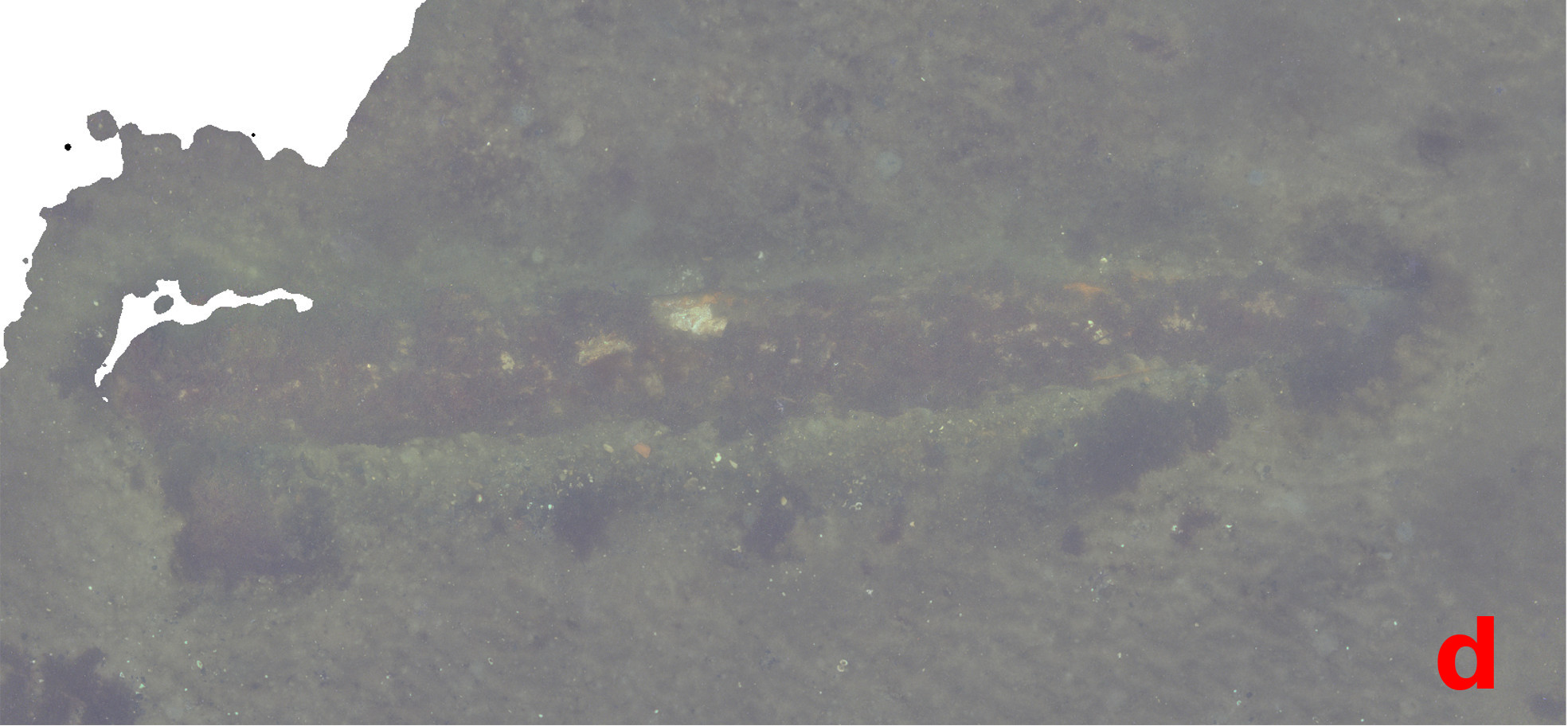}
	\vspace{0.5mm} 
	\includegraphics[width=0.48\columnwidth]{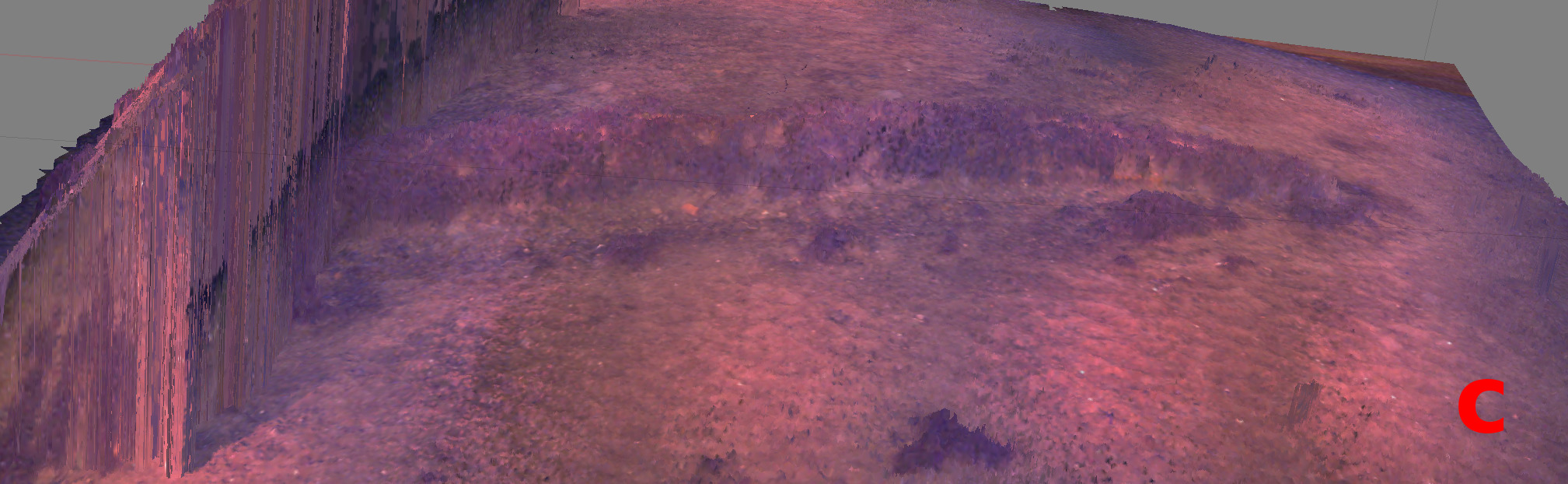}
	\vspace{0.5mm}
	\includegraphics[width=0.48\columnwidth]{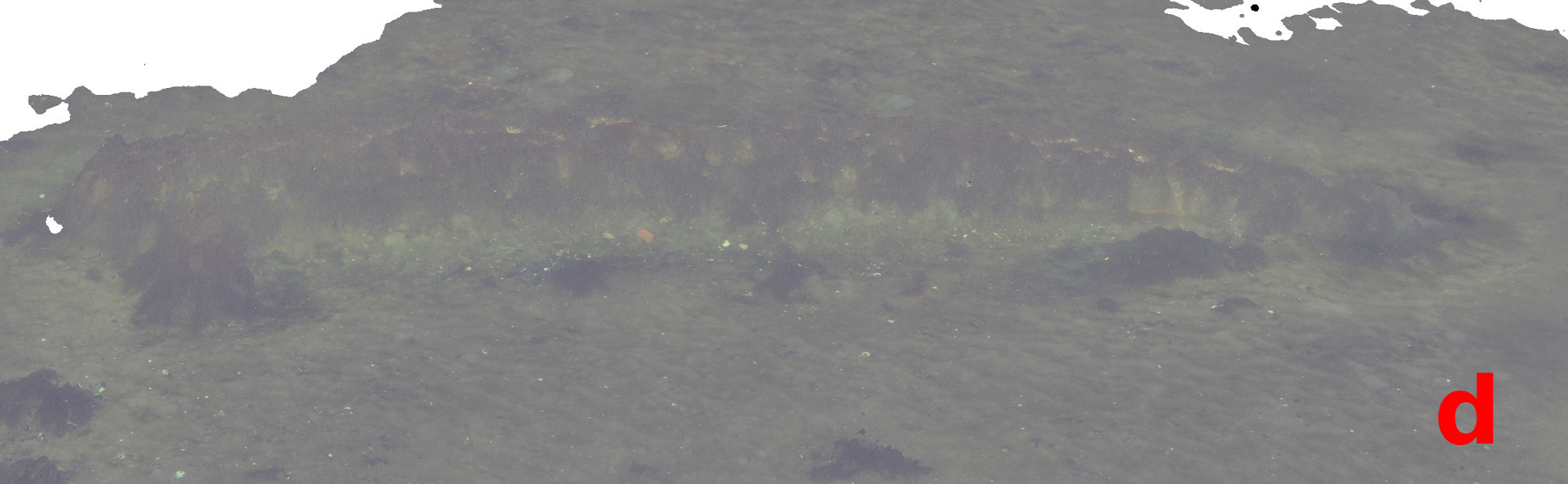}
	\vspace{0.5mm} 
	\includegraphics[width=0.48\columnwidth]{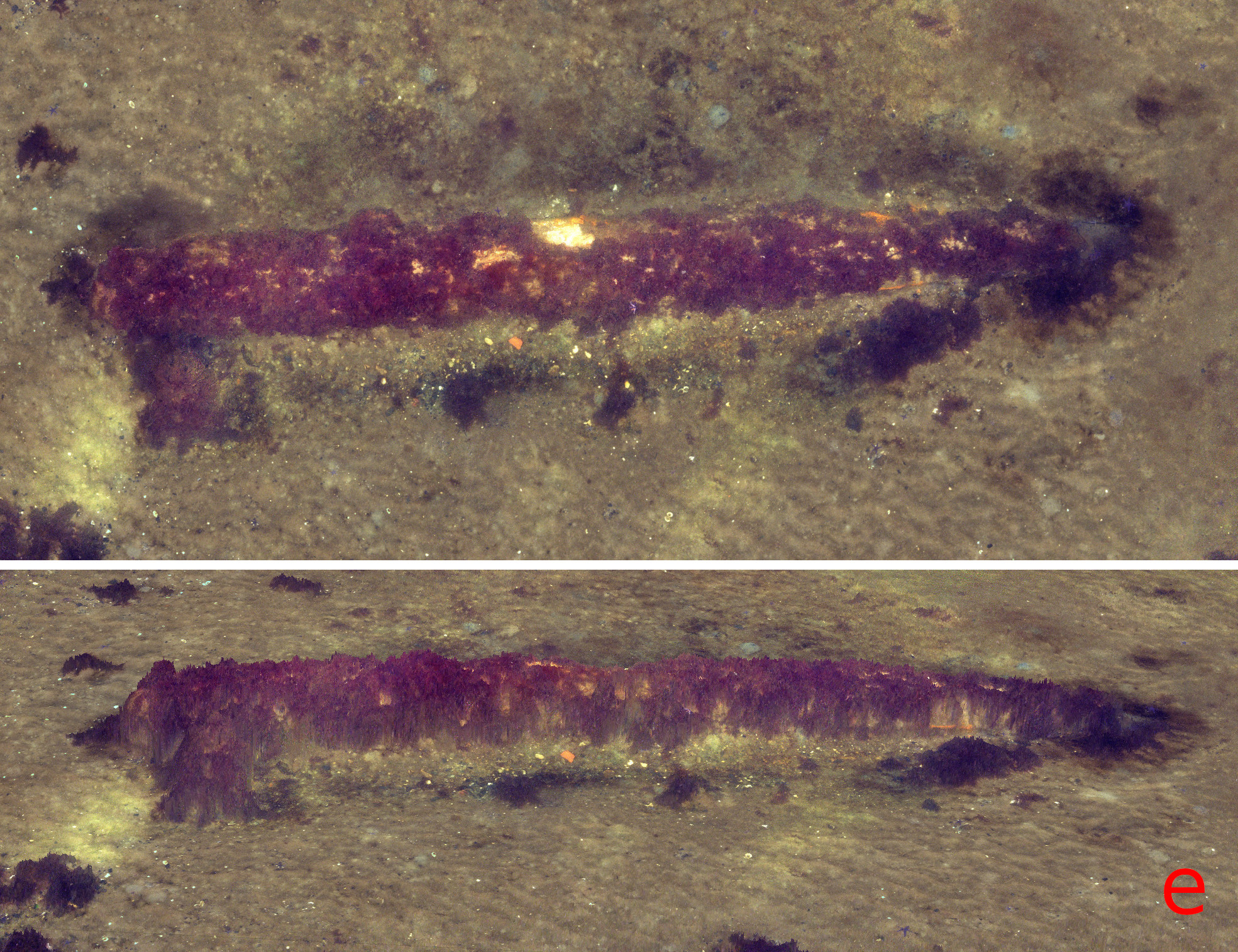}
	\includegraphics[width=0.48\columnwidth]{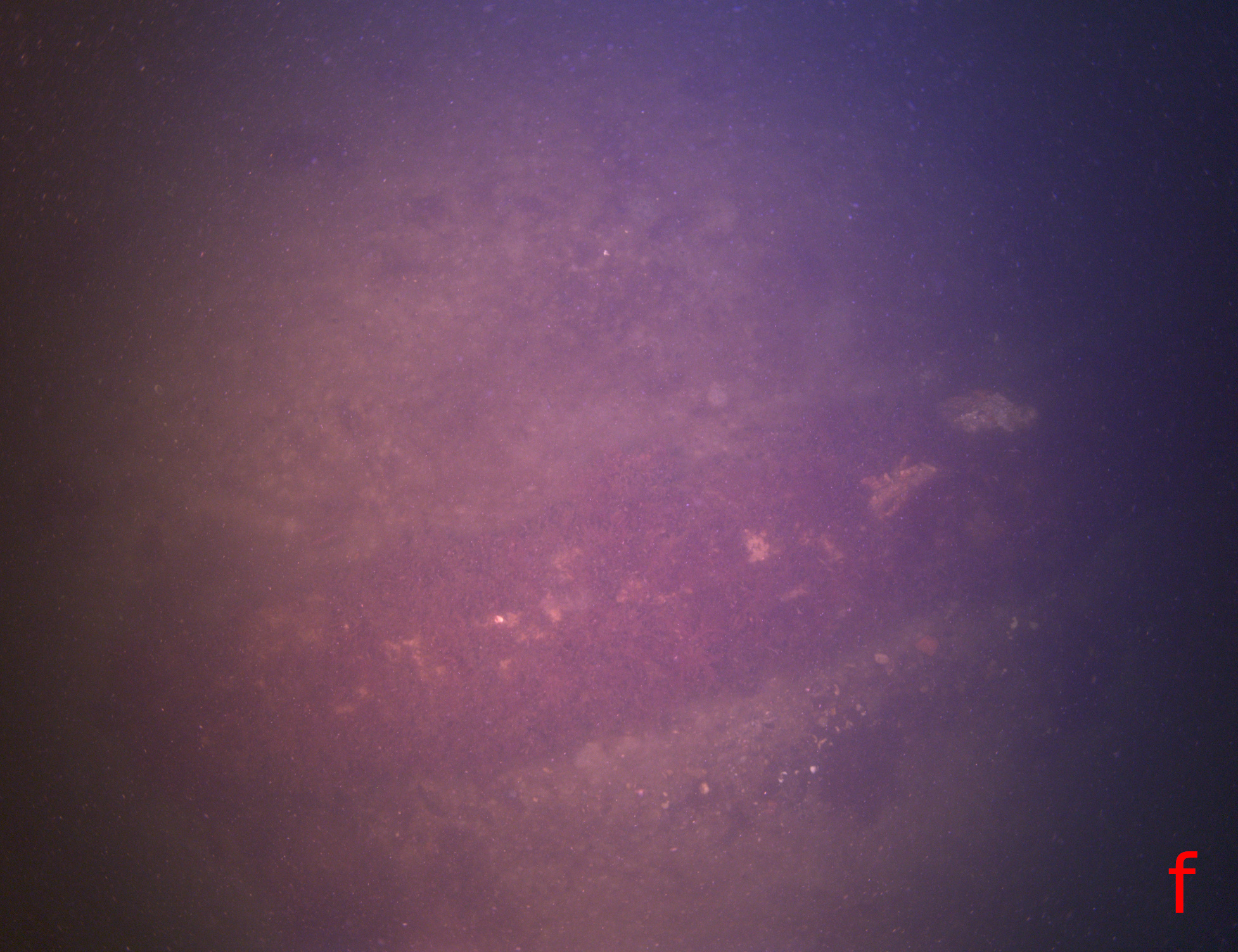}
	\caption{3D reconstruction of a torpedo. The input images have been recorded with a white balance that stresses red too much. In (a) we see the 3D reconstruction using the raw images (each time a top view and a side view). It can be seen that the structure of the torpedo was not recovered correctly. (b) shows the reconstruction from the "Fusion Enhanced"\cite{ancuti12enhancing} images, (c) from "Multi Exposure Fusion"\cite{galdran18dehazingmulti-exposurefusion}, (d) from homomorphic filtering with a fourth order polynomial \cite{Sing_2007-towardsimaging} and (e) from our proposed
	method. Only the reconstruction from our corrected images produces a straight torpedo model and a flat seafloor, whereas the reconstructions from all other image version have problems with the image data.}
	\label{fig:torpedo}
\end{figure}

In fig. \ref{fig:3d} we have run a commercial 3D reconstruction software (Agisoft Photoscan) on three sample images of the flat seafloor. The software will first find correspondences, do robust matching and epipolar geometry estimation, triangulation of correspondences and bundle adjustment, dense estimation and finally produce a 3D model. The top part of the figure shows the raw images and two screenshots of a partially failed, blueish reconstruction. In the bottom part, we have run the same software with the same parameters on the same images, but this time the enhanced version of the images, and we obtain a detailed, consistent 3D model, where we cannot see the seams. In fig. \ref{fig:torpedo} we perform a similar experiment on about 100 images taken by a different AUV with different camera and light system in the Baltic Sea, showing a world war torpedo sunken into the sediment (see figure \ref{fig:torpedo}f for a sample image in this very murky water). Also here we enhance the images prior to 3D reconstruction with our method and other approaches and show the results. Note that the torpedo example is taken in very turbid greenish coastal waters where capturing has to take place from low altitude (ca. 1m), whereas the deep sea data is much clearer and captured from high altitude (ca. 5m).
It should be clear that an entire 3D reconstruction pipeline depends on many parameters and design choices, such that this should just be seen as an example. However, the result is in agreement with our own finding that our images are a useful preprocessing step before matching (both for sparse and dense correspondences).

%\begin{figure}
%	\centering
%	\includegraphics[width=0.48\columnwidth]{torpedo/sample_img002828_20200430_105136519089.jpg} \vspace{0.5mm}
%	\includegraphics[width=0.48\columnwidth]{torpedo/3D_model_topview_proposed_method.jpg} \\
%	\vspace{0.5mm} 
%	\includegraphics[width=0.48\columnwidth]{torpedo/3D_model_sideview_proposed_method.jpg} 
%	\caption{Top: Sample image of torpedo dive using a white balance that strongly amplifies red. Bottom images. 3D reconstruction of a torpedo as in previous figure, but now images pre-normalized with the method proposed in this manuscript. Only the reconstruction from our corrected images produces a straight torpedo model and a flat seafloor, whereas the reconstructions from all other image version have problems with the image data.}
%	\label{fig:torpedo_ours}
%\end{figure}

\begin{figure}
	\centering
	\includegraphics[height=15cm]{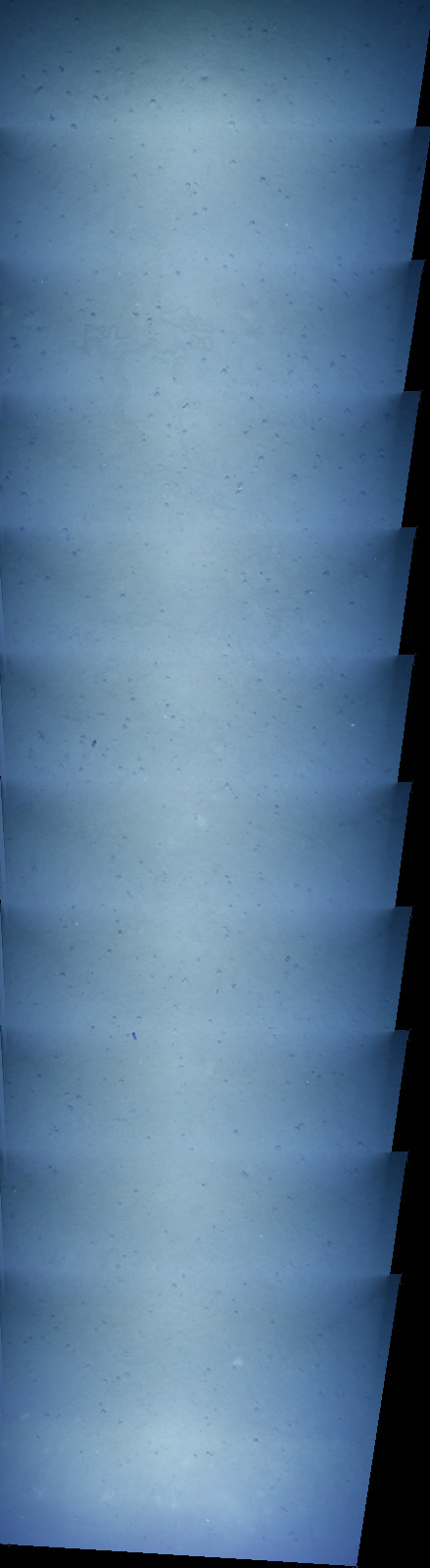} 
	\includegraphics[height=15cm]{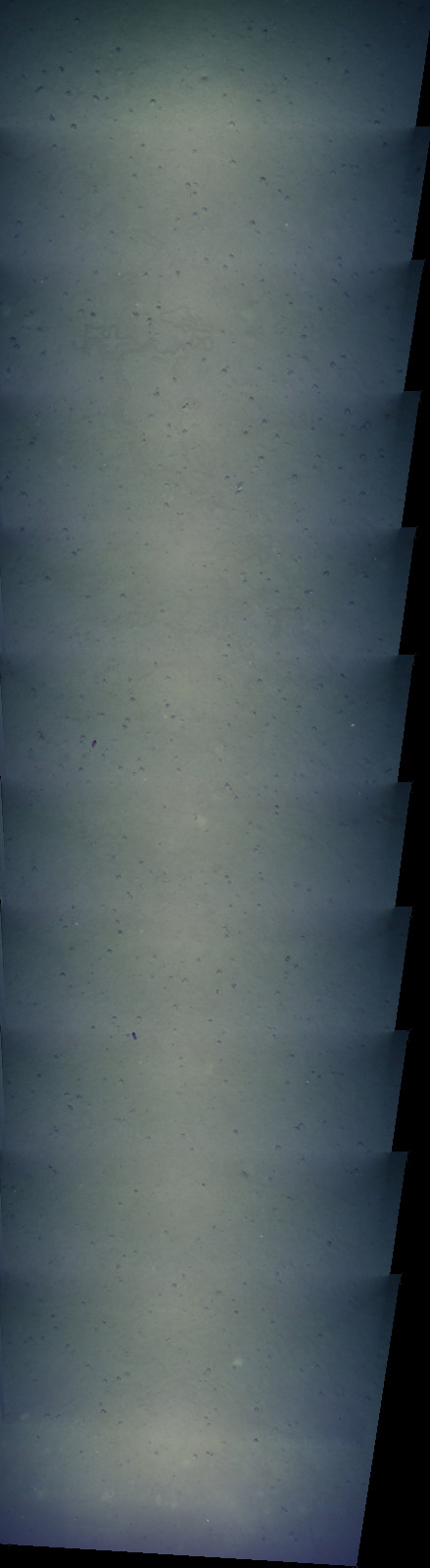}
	\caption{Deep sea mosaic stitched from raw photos (left) and from intermediate images after removing backscatter (right).}
	\label{fig:mosaicunnormalized}
\end{figure}

\begin{figure}
	\centering
	\includegraphics[height=14cm]{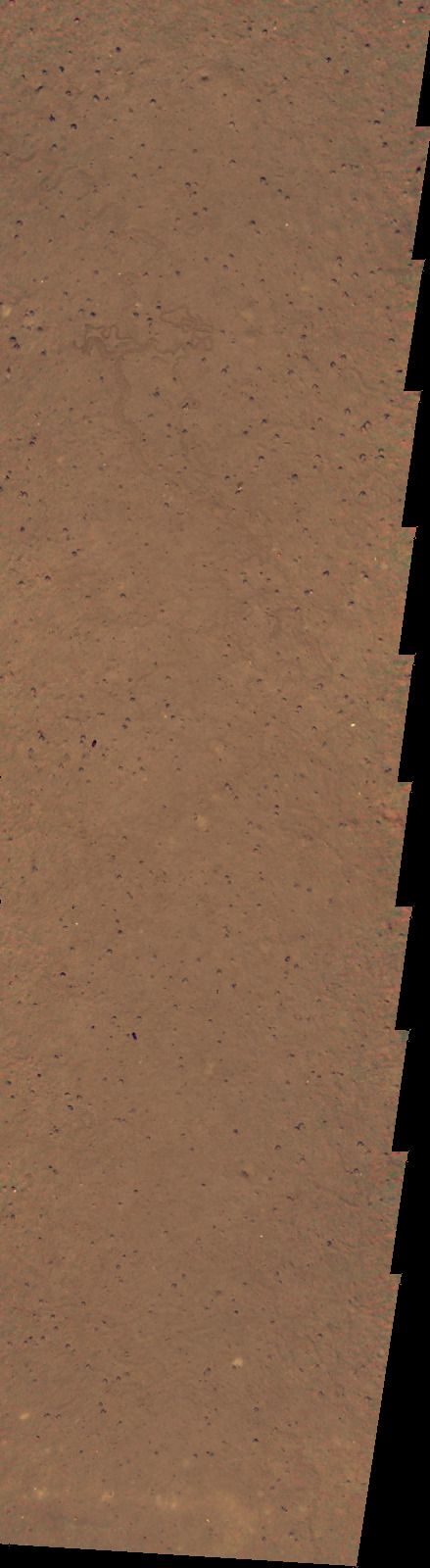}
	\includegraphics[height=14cm]{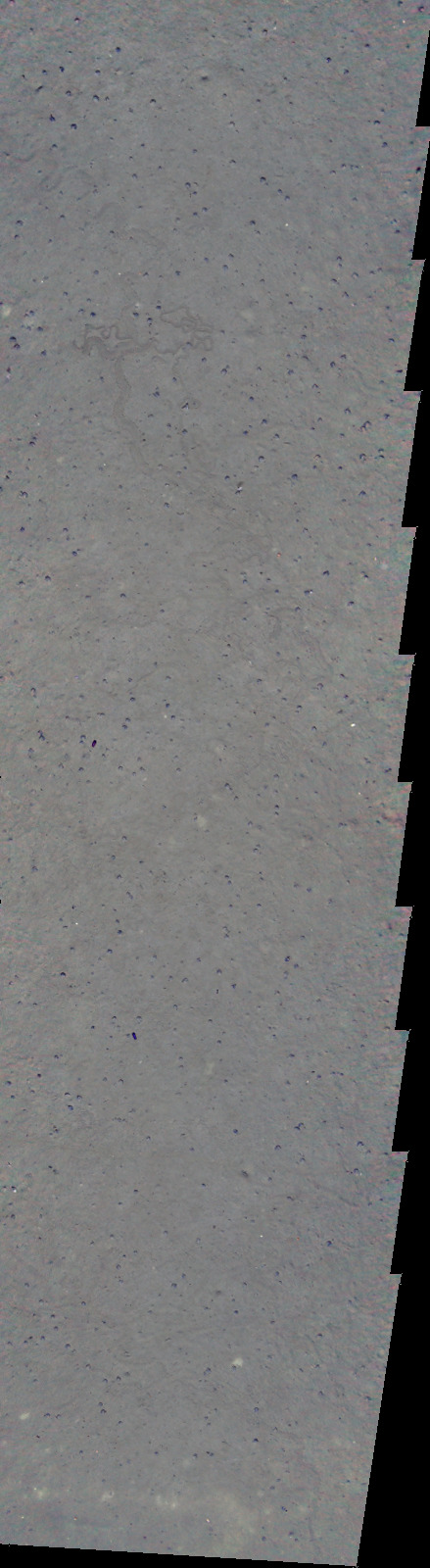}
	\includegraphics[height=14cm]{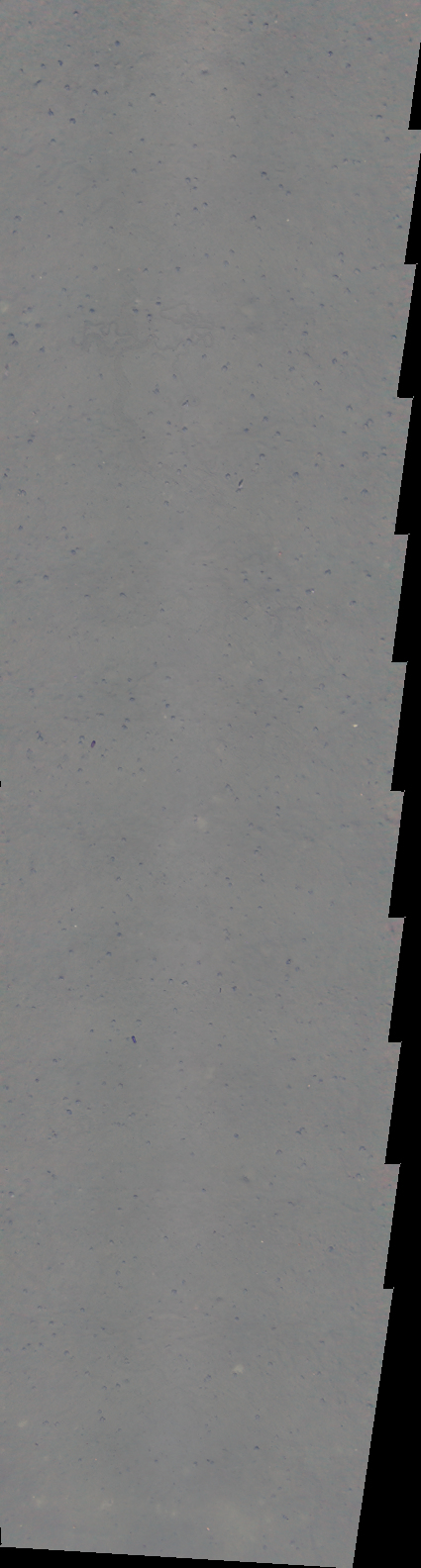}
	\caption{Deep sea mosaic stitched from images after removing illumination, with post-white balance (assumed seafloor color) brown (left), or grey (center). For comparison, we show results using the method of \cite{Sing_2007-towardsimaging} (right), which exibits an overly bright corridor in the center and suffers from low contrast.}
	\label{fig:mosaicnormalized}
\end{figure}

\subsection{Deep Sea Mosaic}
In figures \ref{fig:mosaicunnormalized} and \ref{fig:mosaicnormalized}, we have registered the normalized versions of an image sequence taken at more than 4km water depth in polymetallic nodule fields at the seafloor of the Pacific Ocean\cite{greinert2017siar}. The distance between camera and a 24-LED-flash (4kW electrical power) was about 2m, and the flying altitude was slightly less than 5m, taking one image per second with 1.5m/s speed and an across track field of view of 90$^\circ$ (undistorted). As outlined, the seafloor color can be assumed as grey (fig. \ref{fig:mosaicnormalized}, right) or can be set to the color of a seafloor sample if available (fig. \ref{fig:mosaicnormalized}, left), e.g. when taking samples\cite{greinert2015_SO242}. The micro-navigation of the deep sea robot has been obtained using structure from motion techniques on the enhanced images.  This navigation information is then used to stitch the raw images (fig. \ref{fig:mosaicunnormalized}) and also the enhanced versions (fig. \ref{fig:mosaicnormalized}) of the images. For stitching we use a simple two-band blending with rectangular weight that goes from 1 (image center) to 0 (image boundary): The high frequency information is taken from the pixel with the highest weight, the low frequency information is averaged among the images. It should be clear that the blending can be further tuned and improved, but the goal here is not to hide the problems, but rather to show them. The raw mosaic suffers from strong illumination effects at the right and left boundaries. If the mosaic is created from 10000 images, it will be dominated by the illumination patterns.

\begin{figure}
	\centering
	\includegraphics[height=10cm]{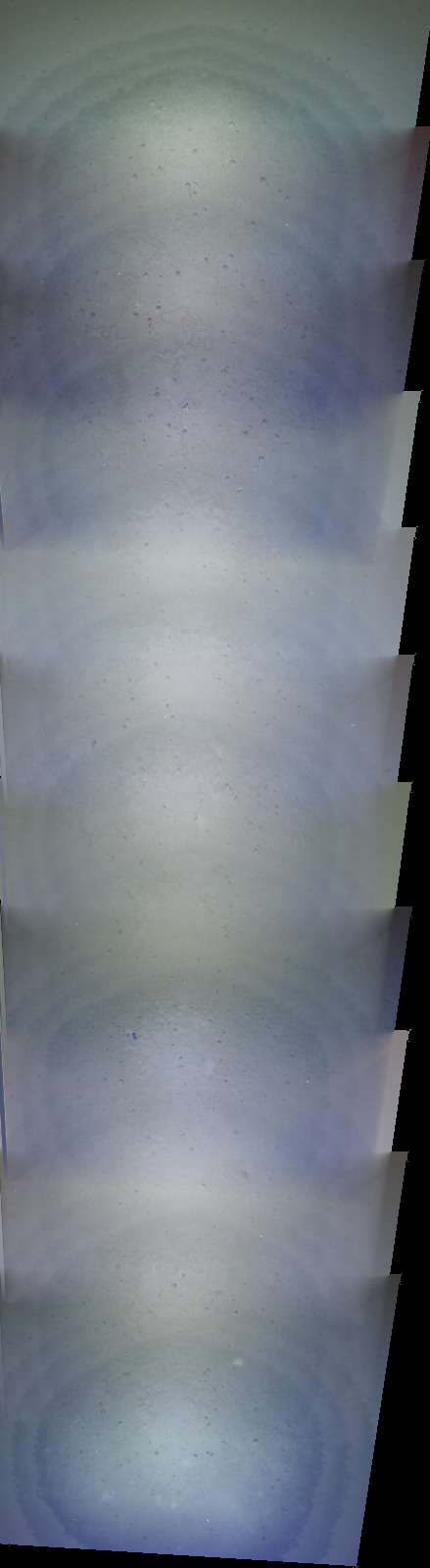}
	\includegraphics[height=10cm]{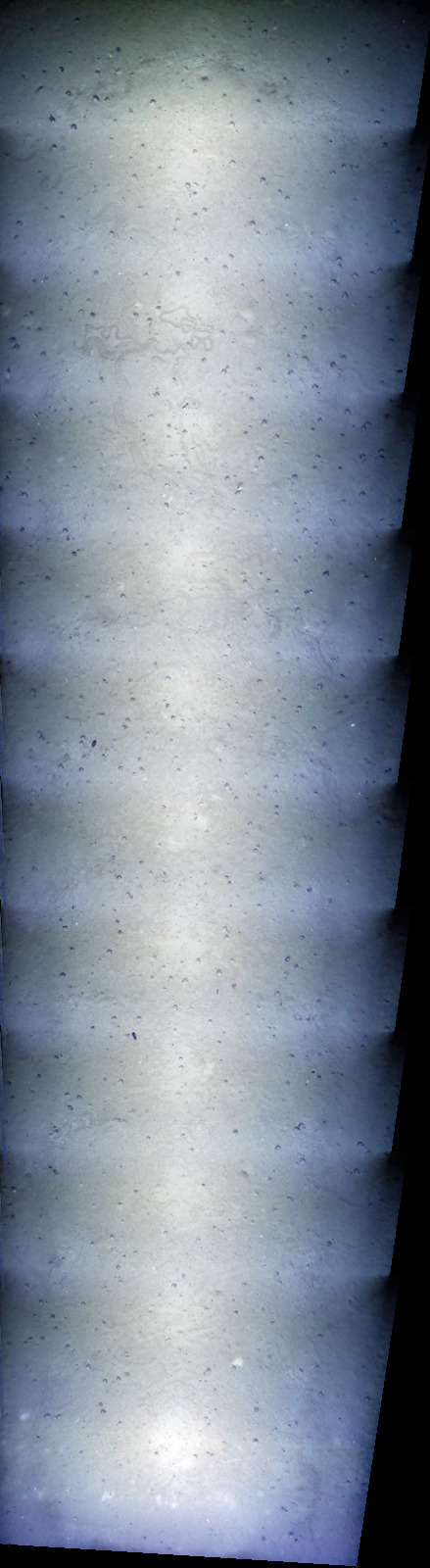}
	\includegraphics[height=10cm]{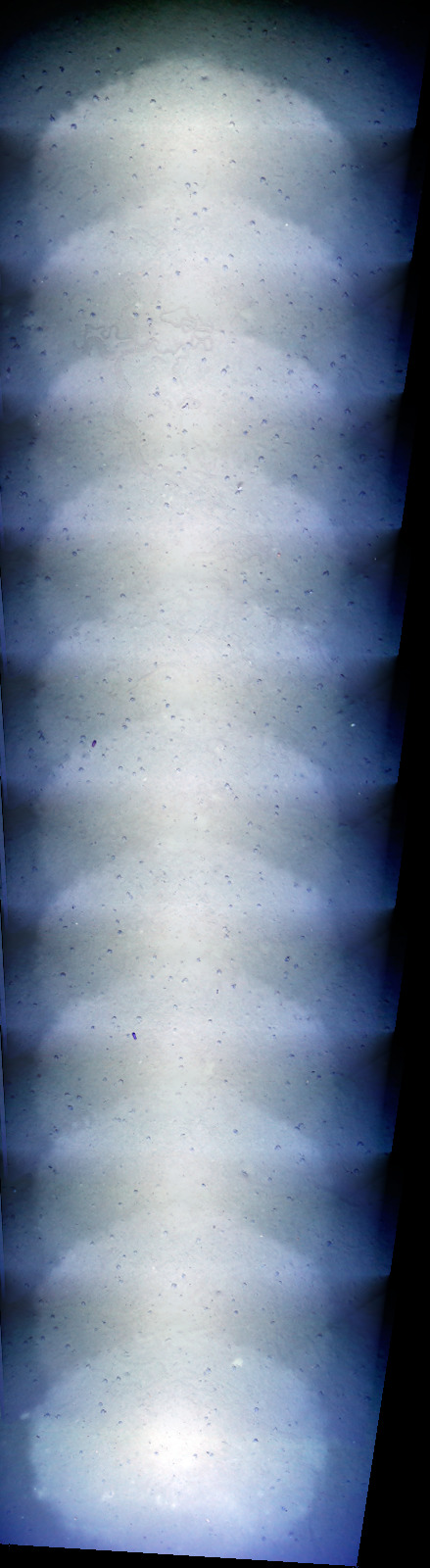}
	\includegraphics[height=10cm]{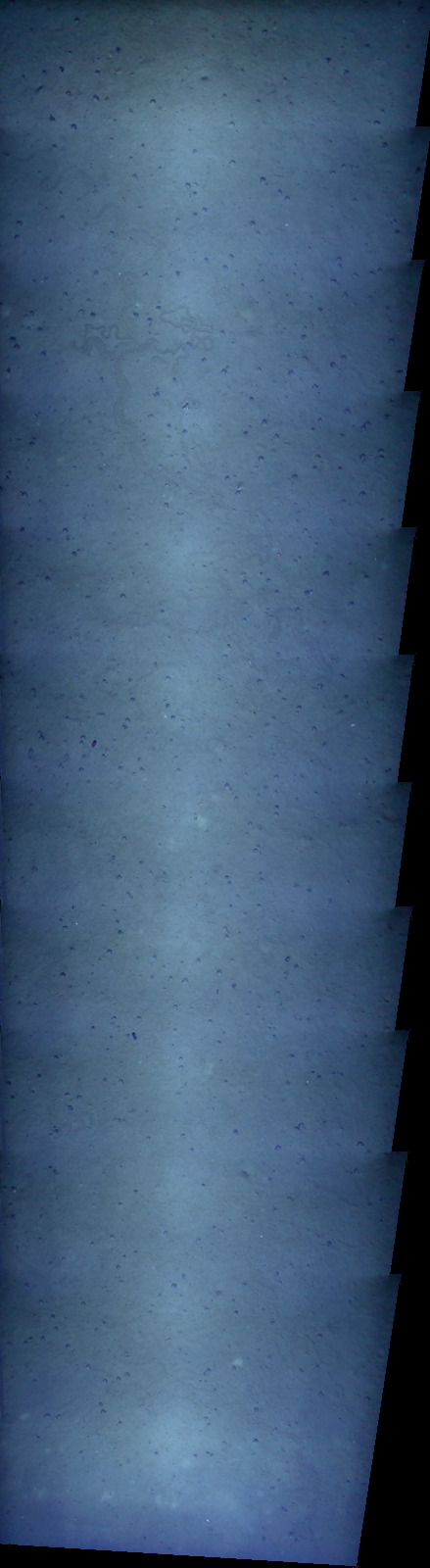}
	\caption{From left to right: Deep sea mosaic stitched using "Optimized Contrast Enhance"\cite{KIM2013410}, "Fusion Enhance"\cite{ancuti12enhancing}, "Remove Backscatter"\cite{zhang2016removing} and "Multi Exposure Fusion"\cite{galdran18dehazingmulti-exposurefusion}. All mosaics are stitched using the same geometry and blending parameters, they only differ in the color preprocessing of the input images. It can be seen that large scale maps using any of these methods (or the raw images) would be dominated by artefacts and lighting effects and it would be very hard to identify the real seafloor structures. In particular, we are only showing one track of images and all approaches produces a brighter stripe in the center. In a lawnmower mission we would see many of these bright artefacts stripes next to each other.}
	\label{fig:mosaiccompetitors}
\end{figure}

\begin{figure}
	\centering
	\includegraphics[width=0.98\columnwidth]{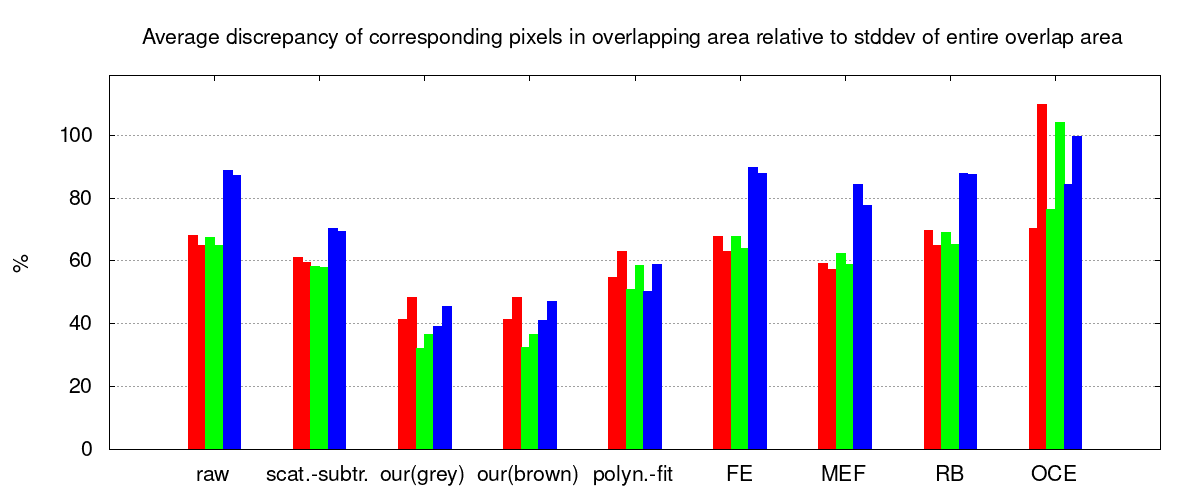}
	\caption{Consistency error in deep sea mosaic stitching. For each of the six approaches (RB \cite{zhang2016removing}, MEF \cite{galdran18dehazingmulti-exposurefusion},OCE \cite{KIM2013410}, FE \cite{ancuti12enhancing}, polynomial fit\cite{Sing_2007-towardsimaging} and ours) we compare the color mapped onto the seafloor orthophoto from the different overlapping input images, in case more than one image covers this area. The images are backward mapped using trilinear interpolation on an image pyramid. We compute the average absolute difference from the mosaic color for each pixel, for each of the R,G,B values of each pixel, in both the upper and lower part of the mosaic, which is why we present two red, two blue and two green bars per method. Afterwards we divide this value by the standard deviation of all pixels in the overlapping area to make the measure insensitive against offsets and scalings. Smaller errors mean better agreement (0 would be perfect agreement and consistent color correction), but some error is inevitable even for perfect restoration, since small sub-pixel misalignments at bright-dark transitions or slight micro-relief shadows increase the error.
		Still, it can be seen qualitatively from the result that our method is the most consistent, the error does not change when using brown or grey white balance and removing scatter already improves the inconsistency a little bit. 
		 Please note that this consistency evaluation even does not penalize the obviously present dark boundary in the other methods, as long as it is consistently dark and the results would be more extreme when using multiple rows of the mosaic.}
	\label{fig:consistency}
\end{figure}
\

We also qualitatively compare our technique to fusion enhancement\cite{ancuti12enhancing}, multi exposure fusion\cite{galdran18dehazingmulti-exposurefusion}, optimized contrast enhancement\cite{KIM2013410} and backscatter removal\cite{zhang2016removing}  using the implementation and default parameters provided in \cite{wang19reviewenhancement}. Note that most of these approaches were not designed for deep sea scenarios, but we believe it is nevertheless interesting to qualitatively see the effects. The results are displayed in fig. \ref{fig:mosaiccompetitors} and it can be seen that none of these methods produces consistent results for the deep sea light cone setting. It would very likely be possible to improve the results by tuning parameters, but it can be seen that all approaches suffer from inconsistency between overlapping images. The key idea of our proposed solution is that we do not need to sit down after each mapping campaign and manually adjust the parameters and retrain the algorithms. This would be impractical and thus we need a parameter-free approach.
The only approach that produced reasonable images is \cite{Sing_2007-towardsimaging} by fitting a 4th order polynomial in log space. However, the approach only considers multiplicative effects, which results in a loss of contrast, and the degree of the polynomial has to be adapted to the light cone: If the polynomial degree is too high, it will fit to (and remove) scene structures, if too small, it cannot cover the illuminatin pattern (e.g. for multi-LED setups).

Please note that for our approach knowledge and calibration of light sources, water properties and camera calibration is not needed, neither 3D reconstruction of the scene, and therefor approaches that require all these parameters cannot be compared (e.g. \cite{bryson16colorcorection}).

\begin{figure}
	\centering
	\includegraphics[width=0.48\columnwidth]{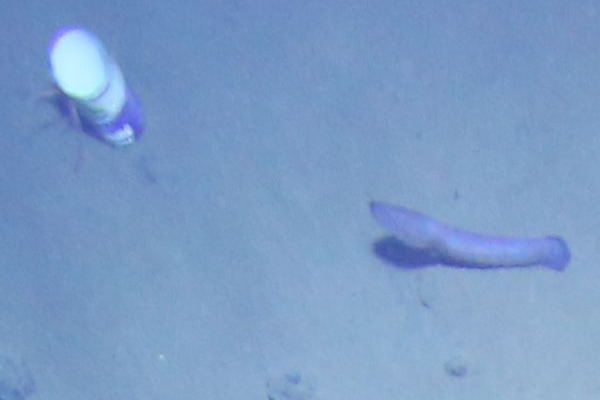} 
	\includegraphics[width=0.48\columnwidth]{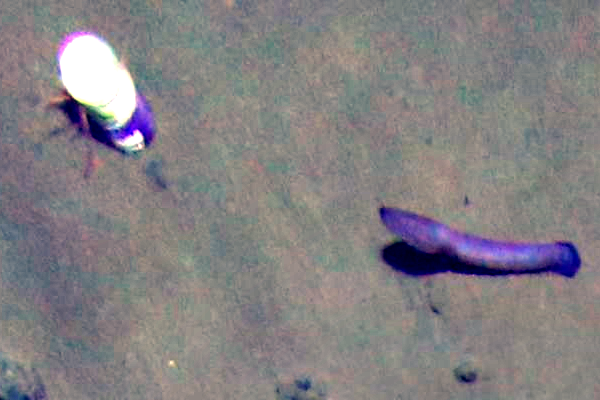} \\
	\includegraphics[width=0.48\columnwidth]{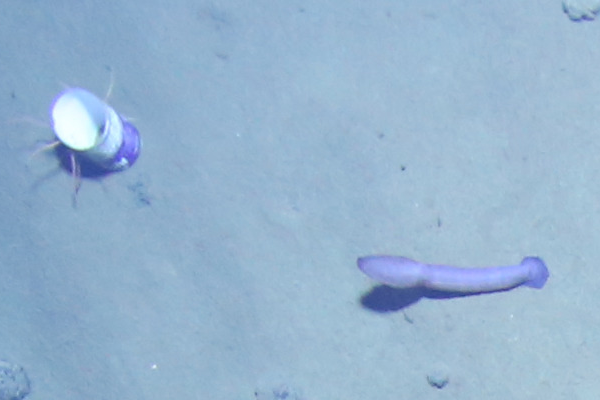} 
	\includegraphics[width=0.48\columnwidth]{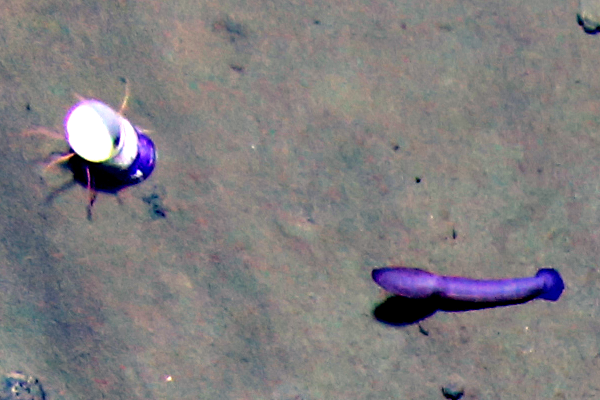} \\
	\includegraphics[width=0.48\columnwidth]{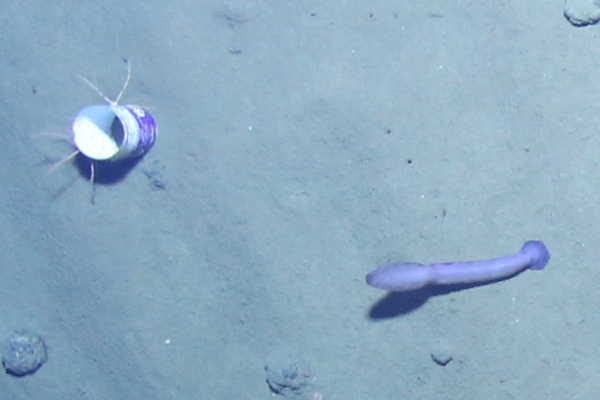} 
	\includegraphics[width=0.48\columnwidth]{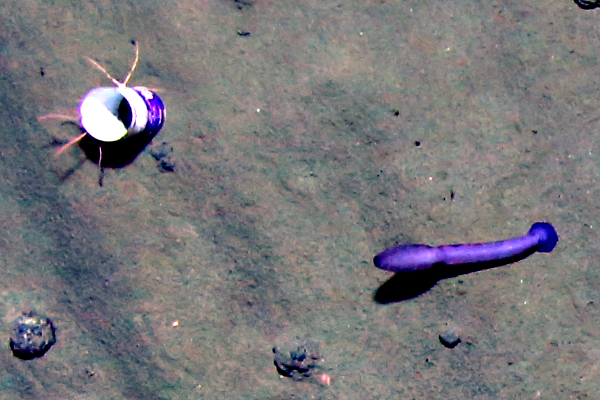} \\
	\includegraphics[width=0.48\columnwidth]{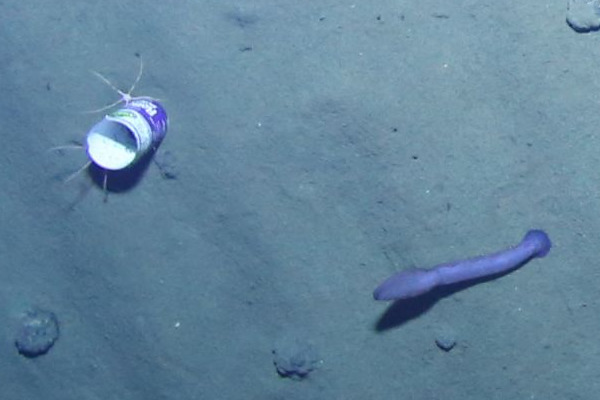} 
	\includegraphics[width=0.48\columnwidth]{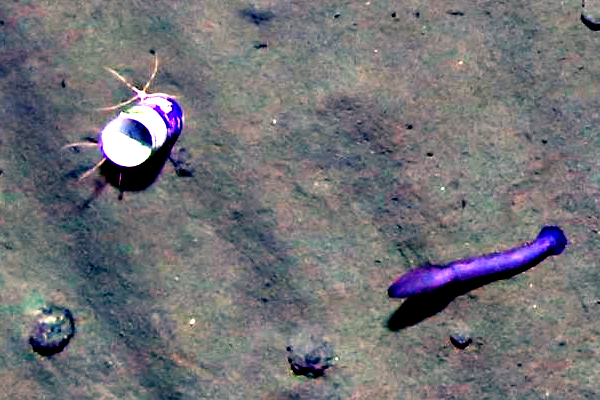} \\
	\caption{Color consistency of photos. Left: raw image cutouts. Right: corresponding enhanced images.}
	\label{fig:colorcomparison}
\end{figure}

\subsection{Consistency}
For the deep sea mosaics in figures \ref{fig:mosaicnormalized},\ref{fig:mosaiccompetitors}, we also numerically analyse how consistent the colors of corresponding seafloor areas are. Before blending, we compare the backward-mapped colors of a seafloor mosaic pixel from different input images and compute their difference to the mosaic. This is averaged over all pixels that are seen from more than one input photo to obtain a RMSE. Finally, this number is divided by the standard deviation of all pixels produced by the respective normalization method, in order to avoid a bias for methods that just make the image dark or uniform color (which would be maximally consistent). Our method for deep sea lighting compensation outperforms all other methods (cf. to fig. \ref{fig:consistency}). 
Note that only our method and \cite{Sing_2007-towardsimaging} do not produce a dark rim towards the image boundaries.
The data set chosen for evaluation does not even fully consider this darkening, as it is just a 1D transect. For large 2D mosaics with multiple tracks next to each other, the consistency margin would be even way higher. Consistency is important for instance for loop detection and to compensate drift, because places seen earlier appear less dissimilar if properly enhanced. It can also be seen, that \cite{Sing_2007-towardsimaging} on the other hand suffers from a loss in contrast, potentially because the model is just multiplicative, but maybe also because the  polynomial can fit to local seafloor structures that are actually no lighting artefact and then overcompensates.

Finally, in fig. \ref{fig:colorcomparison} a cutout from 4 photos of the same object is presented. The left column shows the raw image data and the right column shows the enhanced version. It can be seen that despite quite different raw image appearance, the enhanced result stays qualitatively stable.

\section{Discussion}
When viewing the enhanced images, the illumination patterns are completely removed and objects photographed several times show very consistent colors. Of course, shadows still remain, and shading of micro profiles remains, which makes image matching still challenging. Nevertheless, the method seems well suited as a preprocessing step to remove the largest nuisances, also from old or uncalibrated scenarios and -- as a fast enhancement method -- does a quite good job even for the final map. However, once the enhanced images have helped in registering the images and to find the micronavigation and to reconstruct the surface, one can still run the full parametric image restoration to create maps that are related to a physical model (including measurement uncertainties).

\subsubsection*{Limitations and Failure Cases}
The main assumption of the method is that the seafloor has a constant dominant ($>$50\% of pixels) color and the method enforces this. If the seafloor changes from dark brown to light brown, the algorithm will force this change into the water-light regime. Consequently, the colors of animals or objects at the seafloor would be relatively altered and this situation can only be detected by monitoring the multiplicative image. On the other hand, a change of apparent seafloor color could also be explained by different water composition, leading to different attenuation behaviour. This is a generic problem for all approaches that can only be solved by extra knowledge (e.g. monitoring water properties).

Similarly, when factorizing the image into albedo and lighting uneven illumination {\em patterns} will be removed but there remains an ambiguity about the absolute color unless the seafloor color is (approximately) known. This can be imagined as a constant white balance of the entire mosaic (that can be adapted in post-processing). The effect can be seen in fig. \ref{fig:mosaicunnormalized}. For most mapping applications, the absolute color of the seafloor will be less important than having a consistent map.

\section{Conclusion}
Moving light sources together with attenuation and scattering effects impair mapping in the deep sea, i.e.\ finding correspondences, 3D reconstruction and also making maps without water effects. The nuisances can be decomposed into additive and multiplicative terms and we have presented a robust and automatic method to estimate these terms when the seafloor is predominantly homogeneous and flat. The key observation is that in more than 50\% of the pixels we expect to see a dominant seafloor color, which is also the clear prerequisite for being able to use the algorithm. The method does neither require calibration, nor determining physical water parameters or knowledge about light and camera configuration and therefore has the potential for improving also old seafloor footage or uncalibrated videos from the web. The enhanced images can improve SLAM and 3D reconstruction and can be used as a basis for large scale maps, which traditionally suffer from lighting artefacts that obscure the actual patterns. An efficient implementation allows the algorithm to run in (near) real-time, which is in contrast to other, partially very expensive, correction algorithms.

\bibliography{egbib.bbl}
\label{sec:references}
\end{document}